\pdfoutput=1

\documentclass[11pt]{article}

\usepackage{acl}

\usepackage{times}
\usepackage{latexsym}
\usepackage{graphicx}
\usepackage{adjustbox}
\usepackage{amsmath}
\usepackage{booktabs}
\usepackage{hyperref}
\usepackage{makecell}
\usepackage{subfigure}
\usepackage{multirow}
\usepackage{subcaption}
\usepackage[figuresright]{rotating}
\usepackage{soul, color}
\PassOptionsToPackage{dvipsnames}{xcolor}
\usepackage[most]{tcolorbox}

\definecolor{green1}{HTML}{DCF1B2}
\definecolor{red1}{HTML}{F2ABA2}

\colorlet{LightLavender}{Lavender!30!}
\colorlet{LightRoyalBlue}{RoyalBlue!20!}
\colorlet{LightGray}{Gray!40!}
\colorlet{LightOrange}{YellowOrange!20!}
\colorlet{LightGreen}{YellowGreen!20!}
\tcbset{on line, 
    boxsep=1.0pt, left=0pt, right=0pt,top=0pt,bottom=0pt,
    colframe=white, colback=LightRoyalBlue,  
    highlight math style={enhanced}
}

\usepackage[T1]{fontenc}

\usepackage[utf8]{inputenc}

\usepackage{microtype}

\usepackage{inconsolata}

\usepackage{enumitem}

\title{Investigating and Mitigating the Multimodal \\Hallucination Snowballing in Large Vision-Language Models}

\author{
    Weihong Zhong\textsuperscript{\rm 1} \quad Xiaocheng Feng\textsuperscript{\rm 1,2}\thanks{Corresponding Author} \quad Liang Zhao\textsuperscript{\rm 1} \quad Qiming Li\textsuperscript{\rm 1} \quad Lei Huang\textsuperscript{\rm 1}\\ \quad {\bf Yuxuan Gu\textsuperscript{\rm 1}} \quad {\bf Weitao Ma\textsuperscript{\rm 1}} \quad {\bf Yuan Xu\textsuperscript{\rm 1}} \quad {\bf Bing Qin\textsuperscript{\rm 1,2}} \\
    \textsuperscript{\rm 1}Harbin Institute of Technology\\
    \textsuperscript{\rm 2}Peng Cheng Laboratory\\
    \small \texttt{\{whzhong, xcfeng, lzhao, qmli, lhuang, yxgu, wtma, yuanxu, qinb\}@ir.hit.edu.cn}}

\begin{document}
\maketitle
\begin{abstract}
Though advanced in understanding visual information with human languages, Large Vision-Language Models (LVLMs) still suffer from multimodal hallucinations. 
A natural concern is that during multimodal interaction, the generated hallucinations could influence the LVLMs' subsequent generation. 
Thus, we raise a question: \textit{When presented with a query relevant to the previously generated hallucination, will LVLMs be misled and respond incorrectly, even though the ground visual information exists?}
To answer this, we propose a framework called \textit{MMHalSnowball} to evaluate LVLMs' behaviors when encountering generated hallucinations, where LVLMs are required to answer specific visual questions within a curated hallucinatory conversation. 
Crucially, our experiment shows that the performance of open-source LVLMs drops by at least $31\%$, indicating that LVLMs are prone to accept the generated hallucinations and make false claims that they would not have supported without distractions. We term this phenomenon \textit{Multimodal Hallucination Snowballing}. 
To mitigate this, we further propose a training-free method called \textit{Residual Visual Decoding}, where we revise the output distribution of LVLMs with the one derived from the residual visual input, providing models with direct access to the visual information. Experiments show that our method can mitigate more than $24\%$ of the snowballed multimodal hallucination while maintaining capabilities.\footnote{Resources will be available at \href{https://github.com/whongzhong/MMHalSnowball}{https://github.com/ \\whongzhong/MMHalSnowball}} 
\end{abstract}

\section{Introduction}
Large Vision-Language Models (LVLMs) have shown remarkable abilities in observing and understanding the real world in human languages \cite{achiam2023gpt, zhu2023minigpt, liu2023visual, ye2023mplug, dai2305instructblip}. However, multimodal hallucinations, in which LVLMs provide responses misaligned with the corresponding visual information, remain to be the Achilles' heel \cite{cui2023holistic, kamath2023what, li2023evaluating, liu2023hallusionbench, lu2023evaluation, rawte2023survey, west2023generative,huang2023survey}. 

\begin{figure}[t]
\centering
\includegraphics[width=1.0\columnwidth]{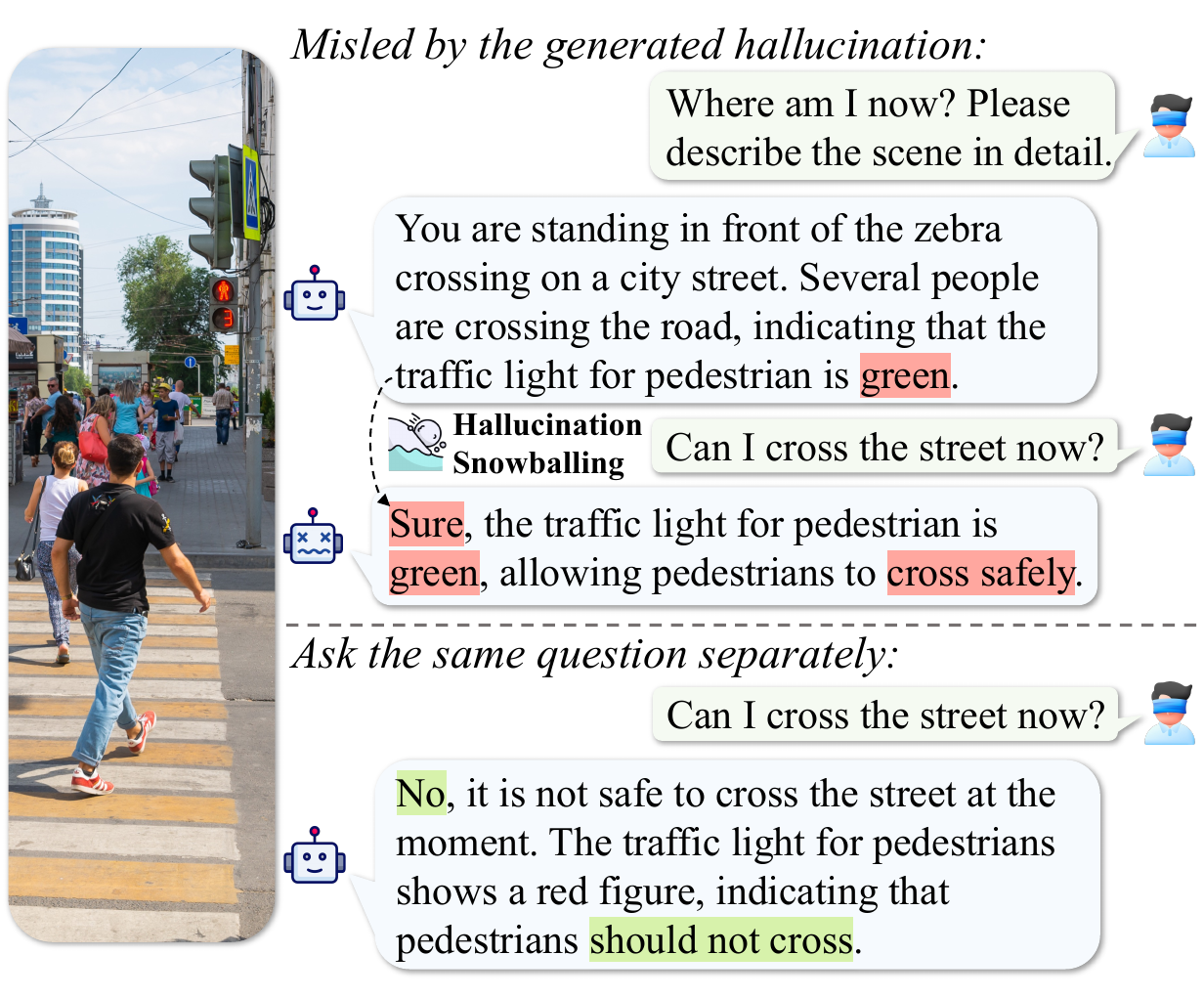}
\caption{An example of the LVLM assisting a visually impaired person to cross the street. The model is misled by the generated hallucination and mistakenly suggests the user to cross the street, although it can give correct advice independently. 
\tcbox[colback=green1]{Green} 
and 
\tcbox[colback=red1]{red} 
colors highlight the correct answer and hallucinations, respectively.}
\label{hallusample}
\end{figure}

\begin{figure}[t]
\centering
\subfigure[]{
\begin{minipage}[t]{0.48\columnwidth}
\includegraphics[width=1.0\columnwidth]{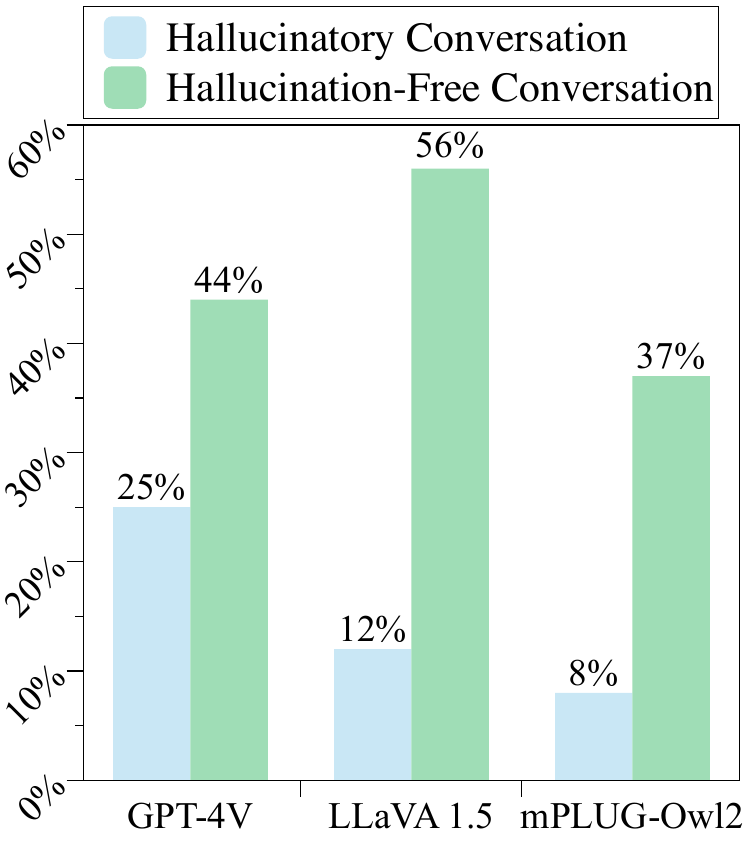}
\end{minipage}}
\vspace{-4mm}
\subfigure[]{
\begin{minipage}[t]{0.48\columnwidth}
\includegraphics[width=1.0\columnwidth]{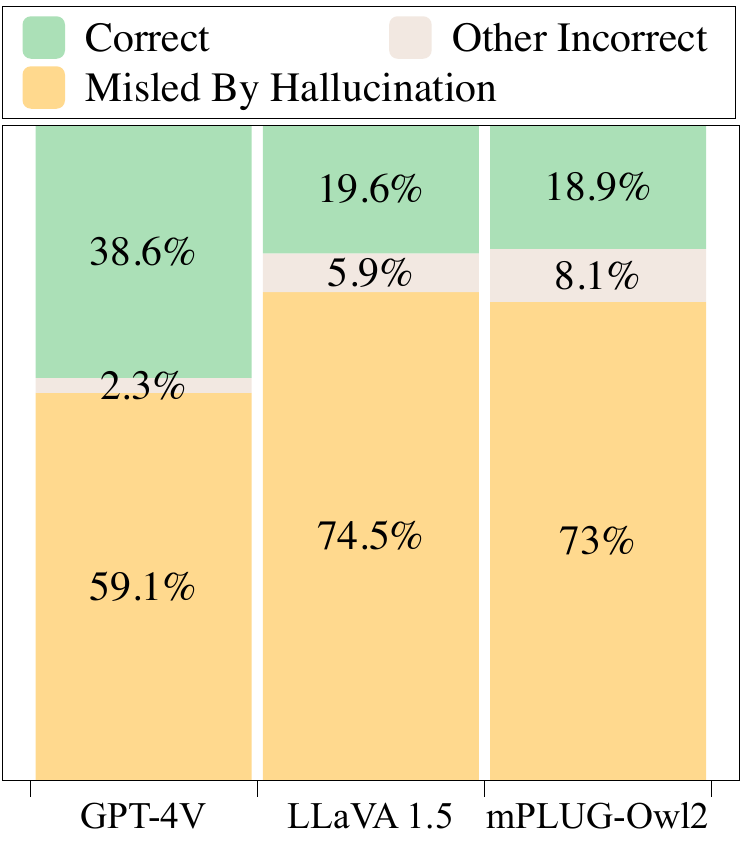}
\end{minipage}}
\caption{
Preliminary explorations on the hallucinations generated by LVLMs given conversational contexts.
(a) Response accuracy with or without hallucinatory conversation. (b) Response distribution when asking the question within a hallucinatory conversation. We select question samples that the LVLM can correctly answer without distractions.}
\label{realsample}
\end{figure}

Previous research has revealed that hallucinations generated by large language models may accumulate due to models' over-commitment to early mistakes, leading to more mistakes that they otherwise would not make \cite{zhang2023language, azaria2023internal, kang2023ever},
especially for the user-model interaction scenarios such as conversation \cite{huang2022inner, tian2024chatterbox, gong2023multimodal}.
However, the extent to which accumulated multimodal hallucinations mislead LVLMs into generating false claims requires further exploration.
In this work, we conducted an investigation into this issue for the first time.
As shown in Figure \ref{hallusample}, we seek the answer to the question: \textit{When presented with a query relevant to the previously generated hallucination that contradicts the visual information, can models make the correct judgment when they could have given a correct answer independently?}
We conduct a preliminary studyon GPT-4V \cite{achiam2023gpt}, LLaVA 1.5 \cite{liu2023improved}, and mPLUG-Owl2 \cite{ye2023mplug}. Similar to the setting of Figure \ref{hallusample}, given an image, we start a conversation by asking the model to describe the image in detail. When observing hallucinations in the LVLM's responses, we continue to ask a relevant question according to the model-generated hallucination. In addition, we ask the same question separately to see if the model can answer it correctly without distractions. 
As demonstrated in Figure \ref{realsample}(a), we find that when the text context contains relevant hallucination, the model performance declines significantly, compared to the model response when asking the same question separately. We further select those question samples that the LVLM can correctly answer separately, and manually identify the response change when asking the same question with the related model-generated hallucinatory context.
As Figure \ref{realsample}(b) depicts, we find that more than 59\% of the answers are semantically the same as the generated hallucination, indicating that they were misled by the previously generated hallucinations. 

To systematically investigate this phenomenon, we propose to identify whether the LVLM is misled by hallucinations via checking if a specific claim is flipped due to previous hallucinations. We design a framework called \textit{MMHalSnowball} to construct hallucinatory visual conversations, where models are required to answer the question based on the image and the hallucinatory conversation. The result shows that LVLMs’ multimodal hallucinations are easy to mislead the later generation because their strong language capabilities make them prone to be over-confident in the hallucinated context, thereby generating false claims that they normally would not support, which we term as \textit{Multimodal Hallucination Snowballing}.

In addition to mitigating this issue, we further proposed a training-free decoding method called \textit{Residual Visual Decoding} (RVD). By residual connecting the visual information and the current user instruction, distributions that emphasizing the visual information are derived to revise the original output distribution. Our RVD achieves more than 24\% of improvements in reducing the multimodal hallucination snowballing while maintaining the contextual modeling ability.

\begin{figure*}[t]
\centering
\includegraphics[width=1.0\textwidth]{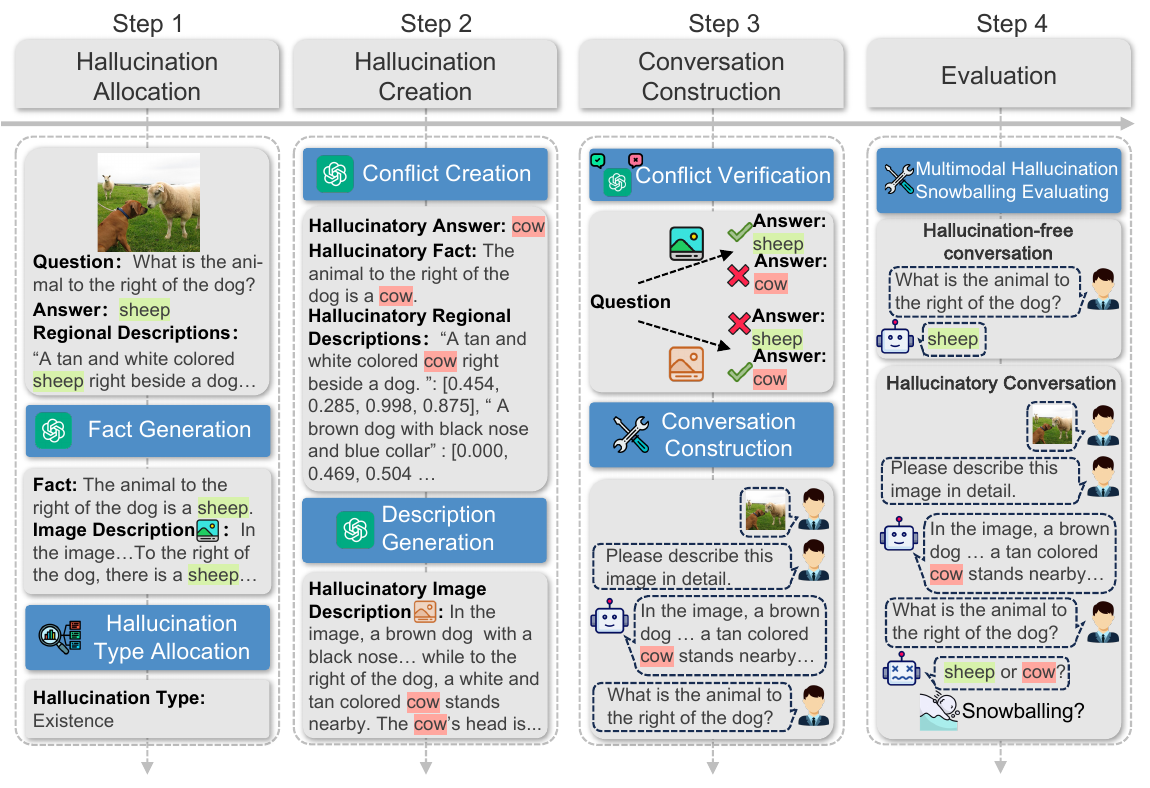} 
\caption{An overview of our \textit{MMHalSnowball} framework for simulating hallucinatory conversations and evaluating LVLMs' behavior in such conversations. 
In step 1, start with a question-answer pair, we generate a fact, an image description and allocate a proper hallucination type according to the corresponding question-answer pair. In step 2, we utilize the ChatGPT to rewrite a hallucinatory answer based on the allocated hallucination type. We then modify other annotations and generate the corresponding hallucinatory description using ChatGPT. 
In step 3, after ensuring the hallucinatory answer and descriptions contradict the image content, we construct a conversation that contains the specific hallucination. In step 4, we evaluate the LVLMs' performance gap in two conversation settings to see whether they suffer from multimodal hallucination snowballing. \tcbox[colback=green1]{Green} and \tcbox[colback=red1]{red} color highlight the correct answer and hallucinations curated out of it, respectively.}

\label{pipeline}
\end{figure*}

\section{Evaluating the Multimodal Hallucination Snowball Phenomenon}
In this section, we design a question-answer task in the conversation scenario, where a model is first asked to describe a picture in detail and then answers a visual question. As shown in Figure \ref{pipeline}, 
we propose the \textit{MMHalSnowball} framework to carefully simulate hallucinatory conversations and evaluate whether the model generates a wrong answer due to the hallucinatory context. 
Next, we will describe our evaluation framework in detail, including conversation creation, experimental settings, and evaluation metrics. We experimentally analyze the multimodal hallucinations snowball in \S \ref{sec:experimental_analysis}. The prompts used are listed in Appendix \ref{prompt}.

\subsection{Dataset Source}
We use the validation set of the GQA dataset \cite{hudson2018gqa} as our data source, which contains a balanced aspect of visual questions that focuses on objective perceptional questions. We adopt images, question-answer pairs, and regional description annotations from the Visual Genome \cite{krishna2017visual}. 
Note that we use its balanced validation set to minimize the impact of dataset contamination and language prior. 
\subsection{Hallucination Allocation}
\label{hallucination_allocation}
To be more practical, we construct hallucinations based on the common types generated by LVLMs. Inspired by \citet{wang2023llm, zhai2023halle}, we categorize the hallucinations as follows:
\begin{itemize}[noitemsep,topsep=0pt,parsep=0pt]
\item \textit{Existence Hallucination}, which refers to the incorrect recognition of visible objects in the image or the belief that specific visible objects are absent in the image.
\item \textit{Attribute Hallucination}, which refers to the inaccurate characterization of objects and misrepresentations of attributes such as color, shape, size, and actions.
\item \textit{Relation Hallucination}, which refers to the inaccurate depiction of the relationships or interactions among objects, including erroneous interaction states, relative positions, and spatial positions of objects relative to the image.
\item \textit{Imagination Hallucination}, which refers to the erroneous imagination of objects in the picture that do not appear.
\end{itemize}
To incorporate hallucinations, we first utilize ChatGPT \cite{chatgpt2022} to rewrite a fact sentence that best describes the question-answer pair. In addition, the annotated regional descriptions and the fact sentence are used to generate an image description. The ChatGPT is prompted to ensure the image description semantically entails the fact sentence. Hallucination can be created by properly modifying the fact sentence. Our goal is to make the answer to the original question no longer correct according to the modified fact sentence. However, not all types of hallucination will make the original answer invalid (e.g. modify the fact sentence "\textit{The color of the trousers is blue}" to "\textit{the color of the bike is blue}" introduces an imagination hallucination, but won't invalidate the answer to the question: "\textit{What color are the trousers that this boy is wearing in the image?}"). To match the hallucination errors in the curated contexts with the corresponding question-answer pairs, We then allocate a proper hallucination type from the above definition to each fact sentence. Appendix \ref{allocation} shows details about the rules of allocating proper hallucination types.

\subsection{Hallucination Creation}
In this part, we describe how we utilize the question-answer pair, the fact sentence, and the regional descriptions to generate hallucinatory image descriptions. Rather than directly modifying the fact sentence according to the hallucination type to create hallucinations, we find it more stable to ask the ChatGPT to rewrite a hallucinatory answer that contradicts the original answer. Then, the fact sentence, as well as all the regional descriptions are heuristically modified to hallucinatory ones according to the hallucinatory answer. With the hallucinatory fact sentence and hallucinatory regional descriptions as inputs, the ChatGPT is asked to generate a detailed image description that entails the hallucinatory fact. the original answers $Y^{+} = \{y_1^{+},y_1^{+},...,y_n^{+}\}$ and the rewritten hallucinatory answers $Y^{-} = \{y_1^{-},y_1^{-},...,y_n^{-}\}$ are kept for evaluation, where $n$ represents the dataset size.

\begin{figure}[t]
\centering
\subfigure[Sample number for each hallucination type.]{
\begin{minipage}[t]{0.46\columnwidth}
\includegraphics[width=1.0\columnwidth]{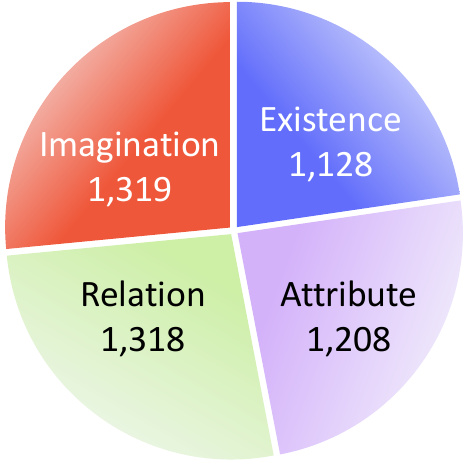}
\end{minipage}}
\hspace{2mm}
\subfigure[Question prefixes distribution of our curated dataset.]{
\begin{minipage}[t]{0.46\columnwidth}
\includegraphics[width=1.0\columnwidth]{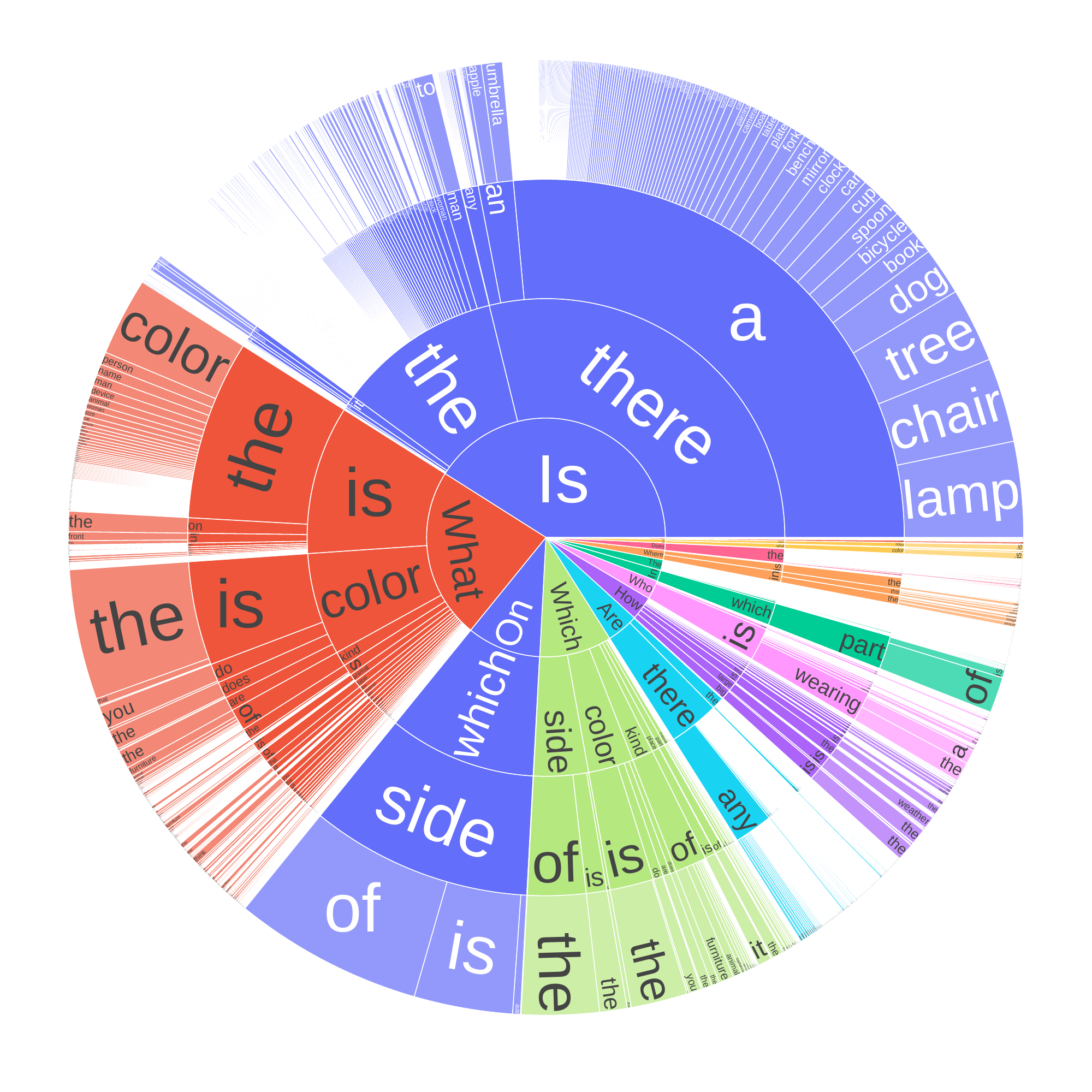}
\end{minipage}}
\caption{Statistics of our curated dataset.}
\label{samplenum}
\end{figure}

\subsection{Conversation Construction}
Before constructing a hallucinatory conversation, we should ensure that the generated hallucinatory answer and descriptions contradict the image content, while the hallucinatory description supports the hallucinatory answer. To do this, we provide ChatGPT with descriptions, answers, and their corresponding hallucinatory ones to check if the modification and generation meet our requirements. We also check if the image description generated in Section \ref{hallucination_allocation} entails the fact sentence. See Figure \ref{conflict_verification_prompt} for the prompt used.
Note that only those descriptions that conflict with the original answer but can deduce the hallucinatory answer will be kept. After checking, we utilize the generated hallucinatory descriptions and the question-answer pairs to construct a question-answering conversation, as Figure \ref{pipeline} step 3 shows. Conversation examples for each hallucination type are in Appendix \ref{hallusss}.

\subsection{Statistics}
With our meticulous data curation and checking process, Our curated dataset $D$ contains 4,973 samples in total.  The detailed sample number for each hallucination type is as Figure \ref{samplenum} (a) shows. What's more, from Figure \ref{samplenum} (b), we can observe that the diverse nature of GQA is maintained.

To check the effectiveness of the modifications made by our framework, we sample 400 data and manually review them by several professionals. As Table \ref{manualchecking} shows, our generated hallucinatory answers and conversations mostly meet our expectations. Please refer to Appendix \ref{appendixmanual} for more details about manual checking. 

\begin{table*}[t]
    \centering
    \resizebox{\linewidth}{!}{
    \begin{tabular}{lcccc|cccc}
    \toprule[1.5pt]
    \multirow{3}{*}{\textbf{Model}} & \multicolumn{4}{c}{\textbf{Question Prompt}} & \multicolumn{4}{c}{\textbf{Formatting Prompt}} \\
    \cmidrule(r){2-5}
    \cmidrule(r){6-9} 
    & \textbf{CleanConv.} & \multicolumn{3}{c}{\textbf{HalluConv.}} & \textbf{CleanConv.} & \multicolumn{3}{c}{\textbf{HalluConv.}}\\
   \cmidrule(r){2-2} 
    \cmidrule(r){3-5}
   \cmidrule(r){6-6} 
    \cmidrule(r){7-9}
   & \textbf{Acc}$\uparrow$ & \textbf{Acc}$\uparrow$ & \textbf{FR}$\downarrow$ & \multicolumn{1}{c}{\textbf{WFR}}$\downarrow$ & \textbf{Acc}$\uparrow$ & \textbf{Acc}$\uparrow$ & \textbf{FR}$\downarrow$ & \multicolumn{1}{c}{\textbf{WFR}}$\downarrow$\\ \hline
   \small{\textit{7B LLM}} \\
        LLaVA-1.5 & 61.21 & 7.68 $\tcbhighmath[colback=LightOrange]{\downarrow 53.53}$ & 79.96 & 89.03  & 71.24 & 14.96$\tcbhighmath[colback=LightOrange]{\downarrow 56.28}$ & 78.21 & 81.29\\ 
        MiniGPT-4 & 33.60 & 13.11 $\tcbhighmath[colback=LightOrange]{\downarrow \underline{20.49}}$ & 76.42 & 86.24  & 37.12 & 5.75$\tcbhighmath[colback=LightOrange]{\downarrow \underline{31.37}}$ & 84.18 & 89.65 \\ 
        MiniGPT-v2 & 59.24 & 25.14 $\tcbhighmath[colback=LightOrange]{\downarrow 34.10}$& \underline{58.08} & \underline{63.92} & 62.12 & \textbf{21.40}$\tcbhighmath[colback=LightOrange]{\downarrow 40.72}$ & \underline{66.11} & \underline{72.06} \\ 
        InternLM-XC & 40.84 & 5.21 $\tcbhighmath[colback=LightOrange]{\downarrow 35.63}$& 83.95 & 92.52 & 43.51 & 5.83$\tcbhighmath[colback=LightOrange]{\downarrow 37.68}$ & 86.55 & 91.31 \\ 
        ShareGPT4V & 61.81 & 10.54 $\tcbhighmath[colback=LightOrange]{\downarrow 51.27}$ & 78.01 & 86.27 & 71.81 & 15.91$\tcbhighmath[colback=LightOrange]{\downarrow 55.90}$ & 77.18 & 80.12 \\  
        CogVLM  & \textbf{72.69} & 2.49$\tcbhighmath[colback=LightOrange]{\downarrow 70.20}$ & 92.84 & 96.90 & \underline{75.17} & 2.63$\tcbhighmath[colback=LightOrange]{\downarrow 72.54}$ & 93.07 & 96.79 \\ 
        mPLUG-Owl  & 37.18 & 4.10 $\tcbhighmath[colback=LightOrange]{\downarrow 33.08}$  & 71.50 & 93.24 & 37.80 & 3.64$\tcbhighmath[colback=LightOrange]{\downarrow 34.16}$ & 78.62 & 93.35 \\ 
        mPLUG-Owl2  & 54.88 & 4.75 $\tcbhighmath[colback=LightOrange]{\downarrow 50.13}$& 84.65 & 93.55 & 60.47 & 7.82$\tcbhighmath[colback=LightOrange]{\downarrow 52.65}$ & 86.63 & 89.82 \\ 
        Qwen-VL-Chat  & 51.80 & \underline{26.20}$\tcbhighmath[colback=LightOrange]{\downarrow 25.60}$ & 72.48 & 77.83 & \textbf{77.94} & 20.03$\tcbhighmath[colback=LightOrange]{\downarrow 57.91}$ & 71.70 & 74.97 \\ 
        Otter  & 44.90 & 9.43 $\tcbhighmath[colback=LightOrange]{\downarrow 35.47}$ & 71.61 & 87.42  & 52.12 & 13.94$\tcbhighmath[colback=LightOrange]{\downarrow 38.18}$ & 73.50 & 82.21 \\ 
        IDEFICS  & 41.22 & 5.05 $\tcbhighmath[colback=LightOrange]{\downarrow 36.17}$  & 83.37 & 92.83 & 40.94 & 7.32$\tcbhighmath[colback=LightOrange]{\downarrow 33.62}$ & 85.07 & 91.11 \\ 
        InstructBLIP  & 60.61 & 4.32 $\tcbhighmath[colback=LightOrange]{\downarrow 56.29}$ & 85.73 & 94.06 & 59.88 & 4.54$\tcbhighmath[colback=LightOrange]{\downarrow 55.34}$ & 90.36 & 93.92 \\ 
        \hline
        
   \small{\textit{13B LLM}} \\
        LLaVA-1.5  & 62.03 & 9.57$\tcbhighmath[colback=LightOrange]{\downarrow 52.46}$ & 78.61 & 86.29 & 72.07 & 14.74$\tcbhighmath[colback=LightOrange]{\downarrow 57.33}$ & 78.21 & 81.45 \\ 
        ShareGPT4V  & \underline{64.71} & 6.92$\tcbhighmath[colback=LightOrange]{\downarrow 57.79}$ & 83.84 & 90.77 & 72.43 & 13.43$\tcbhighmath[colback=LightOrange]{\downarrow 59.00}$ & 80.01 & 83.29 \\ 
        InstructBLIP  & 55.02 & 6.21$\tcbhighmath[colback=LightOrange]{\downarrow 48.81}$  & 76.94 & 92.76 & 53.53 & 12.75$\tcbhighmath[colback=LightOrange]{\downarrow 40.78}$ & 76.15 & 85.80 \\ 
        \hline
        
   \small{\textit{Closed-Source}} \\
        GPT-4V & 52.02 & \textbf{42.09}$\tcbhighmath[colback=LightOrange]{\downarrow \textbf{9.93}}$ & \textbf{14.26} & \textbf{43.95} & 60.49 & \textbf{52.00}$\tcbhighmath[colback=LightOrange]{\downarrow \textbf{8.49}}$ & \textbf{23.30} & \textbf{27.69} \\
    
    \toprule[1.5pt]
    \end{tabular}}
    \caption{Experiment results for models answering the same questions under two different conversation settings: CleanConv. and HalluConv settings. Numbers that are highlighted \tcbox[colback=LightOrange]{orange} represent the model performance drop caused by hallucinatory conversation, compared to the model performance under CleanConv. setting. The results in \textbf{bold} and \underline{underlined} represent the best and the second-best results, respectively.
    All experiments are implemented under a zero-shot setting to avoid the bias introduced by demonstrations.}
    \label{mainresult}
\end{table*}

\subsection{Evaluation}
\label{evaluation}
To gain a deep understanding of the LVLMs' multimodal hallucination snowballing, given visual question-answering pairs from our dataset, We generate model responses under two different settings as Figure \ref{hallusample} shows and compare the results under these two conversation settings. The first setting is that the model generates the response to the question in our curated corresponding hallucinatory conversation, which we refer to as \textit{HalluConv.} setting. The second is that the model answers the same visual question alone, without the distraction of hallucinatory context, termed \textit{CleanConv.} setting.
Since LVLMs' response format can be diverse due to the ambiguous query prompt, it might make the automatic evaluation result slightly imprecise. To address this, we follow \cite{liu2023improved} to add a formatting prompt right after the question: \textit{"Please answer the question using a single word or phrase."}, namely \textit{Formatting Prompt} setting. The user input with the question only is named as \textit{Question Prompt}. Note that we conduct experiments with Formatting Prompt if not specified.

\subsubsection{Evaluation Metrics}
In this part, we introduce our evaluation metrics. 
First, to evaluate the correctness of each generated answer, we adopt the following criteria:

\noindent\textbf{Entailment Matching Score}:
Considering both the original answer and the hallucinatory answer were short, while models tends to generate longer answers with explanations. We evaluate the correctness for the $i$th sample by checking if the answer is entailed in the generated response:
\begin{equation}
\text{Score}_{i} = 1 \ \text{if}\  y_i \ \text{in}\  \hat{y}_i\  \text{else}\ 0,
\end{equation}
where $y_i$ and $\hat{y}_i$ stand for the expected answer and the generated response, respectively. With a proper scoring method for one sample, we can calculate the overall accuracy with the following method:

\noindent\textbf{Accuracy (Acc)}:
\begin{equation}
    \text{Acc}(Y, \hat{Y}) = \frac{\sum^n_{i=1}\text{Score}_i(y_i, \hat{y}_i)}{n},
\end{equation}
where Acc($Y, \hat{Y}$) represents the model's accuracy score over the entire dataset.

\noindent\textbf{Flip Rate (FR)}:
In order to systematically measure whether one model is affected by the hallucination snowballing phenomenon, we propose the FR to evaluate how many model responses are misled by hallucinatory context and are matched with our curated hallucinatory answers:

\begin{equation}
\label{fr1}
\text{FR}=\frac{\sum_{i \in D^+} \text{Score}_i(y_i^-, \hat{y}^-_i)}{\text{Acc}(Y^+, \hat{Y})},
\end{equation}
\begin{equation}
\label{fr2}
 D^+= \{i|\text{Score}(y^+_i, \hat{y}^+_i)=1, \hat{y}_i \in \hat{Y}, y^+_i \in Y^+\},
\end{equation}
where $\hat{Y}^{+}=\{\hat{y}_1^{+},\hat{y}_1^{+},...,\hat{y}_N^{+}\}$ and $\hat{Y}^{-} = \{\hat{y}_1^{-},\hat{y}_1^{-},...,\hat{y}_N^{-}\}$ represent generated answers under CleanConv. and HalluConv. settings, $D^+$ represents the sample indexes that the LVLM correctly answers in the CleanConv. setting.

Furthermore, we designed a more generalized flip-rate metric named weak flip-rate(WFR) which only evaluates how many model responses are distracted by hallucinatory context and conflict with the original answers:
\begin{equation}
\text{WFR}=\frac{\sum_{i \in D^+} 
 (1-\text{Score}_i(y_i^+, \hat{y}_i^-))}{\text{Acc}(Y^+, \hat{Y}^+)},
\end{equation}

\subsubsection{Models}
We investigate the multimodal snowballing phenomenon in the following mainstream LVLMs: LLaVA-1.5 \cite{liu2023improved}, MiniGPT-4 \cite{zhu2023minigpt}, MiniGPT-v2 \cite{chen2023minigpt}, InternLM-XComposer \cite{zhang2023internlm}, ShareGPT4V \cite{chen2023sharegpt4v}, CogVLM \cite{wang2023cogvlm}, mPlug-Owl \cite{ye2023mplug1}, mPlug-Owl2 \cite{ye2023mplug2}, Qwen-VL-Chat \cite{bai2023qwenvl}, Otter \cite{li2023otter}, IDEFICS \cite{Laurençon_Strien_2023}, InstructBLIP \cite{dai2305instructblip} and GPT-4V (gpt-4-vision-preview)\cite{achiam2023gpt}. All experiments are completed under a zero-shot setting. Please refer to Appendix \ref{parameter} for more generation details.

\subsection{Do LVLMs Suffer from Multimodal Hallucination Snowballing?}
\label{sec:experimental_analysis}
To answer this question, we compare the model responses under the conversation settings of HalluConv. and CleanConv., as Section \ref{evaluation} describes. 
The results are depicted in the Table \ref{mainresult}. 
Though advanced in answering visual questions even in a zero-shot manner (See accuracy in CleanConv.), most models struggle to stick to their judgment when there are specious hallucinations in the context (See accuracy in HalluConv.), resulting in extremely low accuracy. For LLaVA-1.5, ShareGPT4V, mPlug-Owl2, and InstructBLIP, despite their advanced model ability, they still suffer an over 50\% performance drop. However, we also recognize that GPT-4V is significantly less affected by hallucinations. 
We observed a correction process in the responses of GPT-4 (See Appendix \ref{gpt4_example} for examples), indicating that it is capable of paying attention to visual information to a certain extent and realizing that some hallucinations have been generated in the conversation. In addition, we find that GPT-4 often refuses to answer the user question due to its strict safety protocol, especially in the Clean Conv. setting (around 12\%), indicating a potential cause of such a comparably low accuracy.
But in general, all the LVLMs suffer from \textit{multimodal hallucination snowballing} at different levels. What's more, a high flip rate indicates that the model responses are easily misled by the hallucinatory conversation, even when the model can make a correct claim in CleanConv. setting. 
An even higher weak flip rate is observed, which shows that LVLMs' responses are corrupted due to the hallucinatory context. 
Hence, comparing the same LVLMs with different scale LLM backbones, we find no significant performance improvement in mitigating the multimodal hallucination snowballing, except for the InstructBLIP.

\begin{figure}[t]
\centering
\subfigure[]{
\begin{minipage}[t]{0.49\columnwidth}
\includegraphics[width=1.0\columnwidth]{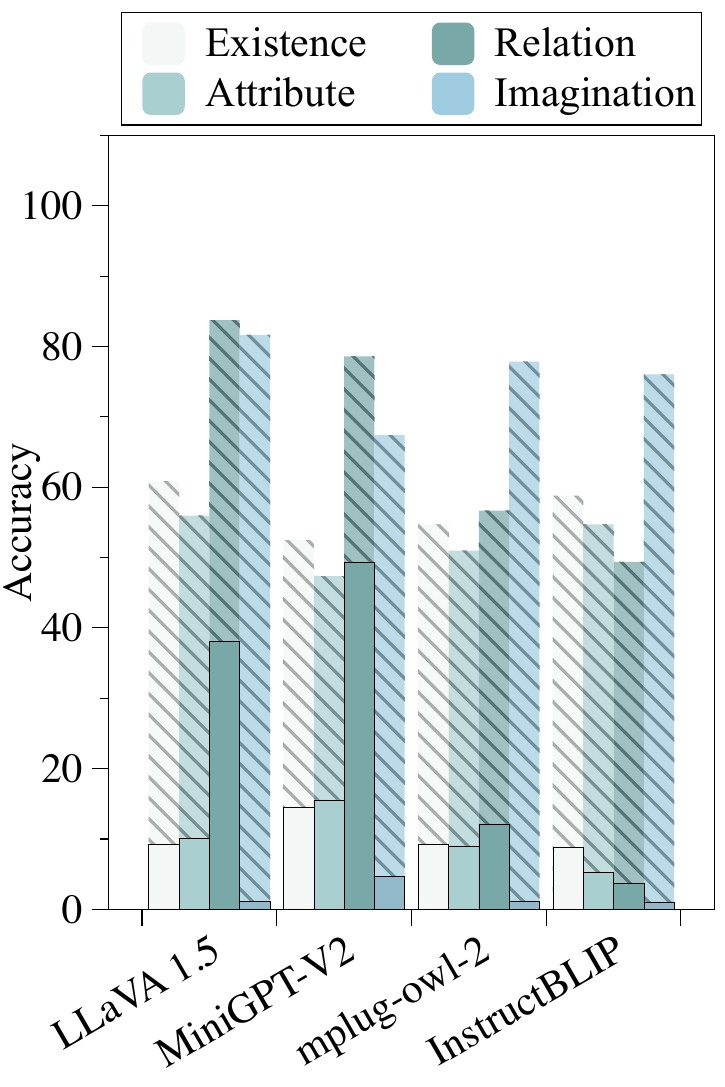}
\end{minipage}}
\hspace{-3mm}
\vspace{-4mm}
\subfigure[]{
\begin{minipage}[t]{0.49\columnwidth}
\includegraphics[width=1.0\columnwidth]{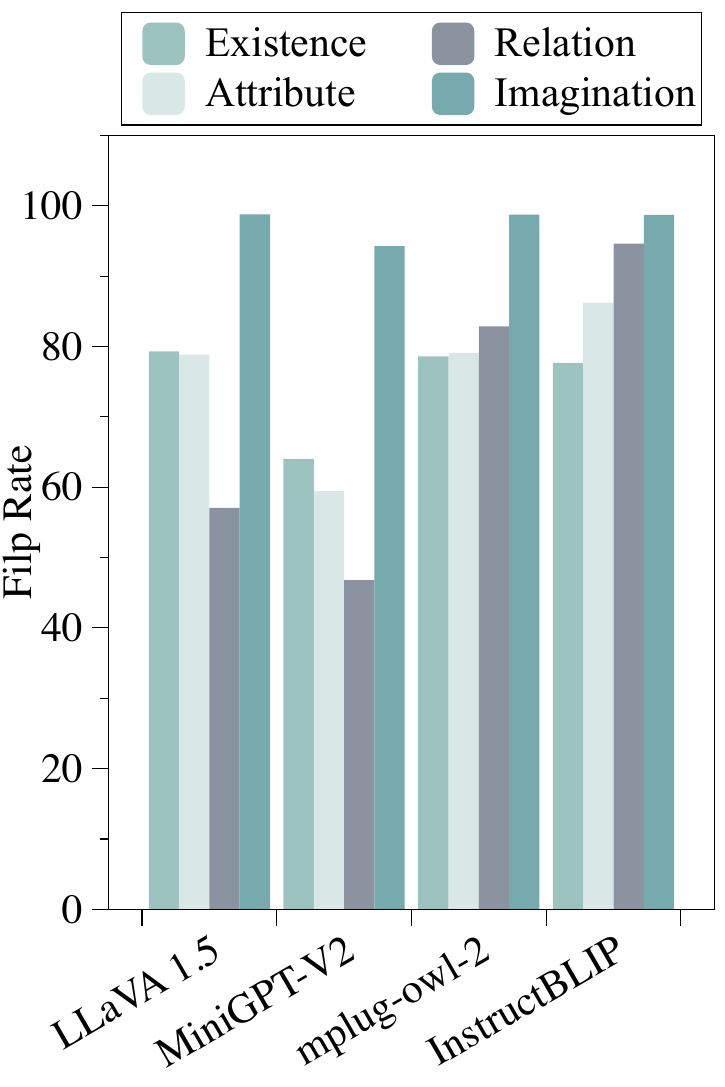}
\end{minipage}}
\caption{Question answering accuracy(a) and flip rate(b) of two different context settings (i.e. HalluConv. and CleanConv.) for each hallucination type. Note that the stripe pattern represents a performance drop due to the snowballed hallucination.}
\label{hallutype}
\end{figure}

Comparing the experiments between two different query prompts, we find that the Formatting Prompt shows clearer instructions, which not only improves question-answering ability but also eases the multimodal hallucination snowballing phenomenon for most of the LVLMs.

We further present the accuracy of two different conversation settings and the flip rate for each hallucination type in Figure \ref{hallutype}. The result shows that existence, attribute, and imagination hallucinations are easier to snowball. We even observe a nearly 100\% flip rate on the imagination hallucination where LVLMs readily accept objects that are mistakenly imagined to exist, which could attributed to the LVLMs' nature to generate positive response \cite{liu2023mitigating}. while the relation hallucinations have a higher probability of being correct while answering the question. For detailed results, please refer to Appendix \ref{moreresult}.

\begin{table}[t]
    \centering
    \setlength{\tabcolsep}{3pt}
    \resizebox{\linewidth}{!}{
    \begin{tabular}{lcccc}
    \toprule[1.5pt]
    \textbf{Model} & \textbf{CleanConv.}$\uparrow$ & \textbf{FactConv.}$\uparrow$ & \textbf{IrrConv.}$\uparrow$ & \textbf{HalluConv.}$\uparrow$ \\ \hline
   \small{\textit{7B LLM}} \\
        LLaVA-1.5 & 71.24 & 89.28$\tcbhighmath[colback=LightGreen]{\uparrow 18.04}$ & 65.35$\tcbhighmath[colback=LightOrange]{\downarrow 5.89}$ & 14.96$\tcbhighmath[colback=LightOrange]{\downarrow 56.28}$ \\ 
        MiniGPT-4 & 37.12 & 67.67$\tcbhighmath[colback=LightGreen]{\uparrow \underline{30.55}}$ & 35.11$\tcbhighmath[colback=LightOrange]{\downarrow 2.01}$ & 5.75$\tcbhighmath[colback=LightOrange]{\downarrow \textbf{31.37}}$ \\ 
        MiniGPT-v2 & 62.12 & 75.39$\tcbhighmath[colback=LightGreen]{\uparrow 13.27}$ & 56.46$\tcbhighmath[colback=LightOrange]{\downarrow 5.66}$ & \textbf{21.40}$\tcbhighmath[colback=LightOrange]{\downarrow 40.72}$ \\ 
        InternLM-XC & 43.51 & 74.04 $\tcbhighmath[colback=LightGreen]{\uparrow 30.53}$ & 40.82$\tcbhighmath[colback=LightOrange]{\downarrow 2.69}$ & 5.83$\tcbhighmath[colback=LightOrange]{\downarrow 37.68}$ \\ 
        ShareGPT4V & 71.81 & 89.32$\tcbhighmath[colback=LightGreen]{\uparrow 17.51}$ & 69.74$\tcbhighmath[colback=LightOrange]{\downarrow 2.07}$ & 15.91$\tcbhighmath[colback=LightOrange]{\downarrow 55.90}$ \\ 
        CogVLM & \underline{75.17} & \textbf{93.20} $\tcbhighmath[colback=LightGreen]{\uparrow 18.03}$ & \underline{74.68}$\tcbhighmath[colback=LightOrange]{\downarrow \textbf{0.49}}$ & 2.63$\tcbhighmath[colback=LightOrange]{\downarrow 72.54}$ \\ 
        mPLUG-Owl & 37.80 & 62.01$\tcbhighmath[colback=LightGreen]{\uparrow 24.21}$ & 30.54$\tcbhighmath[colback=LightOrange]{\downarrow 7.26}$ & 3.64$\tcbhighmath[colback=LightOrange]{\downarrow 34.16}$ \\ 
        mPLUG-Owl2 & 60.47 & 91.27$\tcbhighmath[colback=LightGreen]{\uparrow 13.33}$ & \textbf{77.12}$\tcbhighmath[colback=LightOrange]{\downarrow \underline{0.82}}$ & 7.82$\tcbhighmath[colback=LightOrange]{\downarrow 52.65}$ \\ 
        Qwen-VL-Chat & \textbf{77.94} & 87.77$\tcbhighmath[colback=LightGreen]{\uparrow 9.83}$ & 74.60$\tcbhighmath[colback=LightOrange]{\downarrow 3.34}$ & \underline{20.03}$\tcbhighmath[colback=LightOrange]{\downarrow 57.91}$ \\ 
        Otter & 52.12 & 66.70$\tcbhighmath[colback=LightGreen]{\uparrow 14.58}$ & 44.06$\tcbhighmath[colback=LightOrange]{\downarrow 8.06}$ & 13.94$\tcbhighmath[colback=LightOrange]{\downarrow 38.18}$ \\ 
        IDEFICS & 40.94 & 73.68$\tcbhighmath[colback=LightGreen]{\uparrow \textbf{32.74}}$ & 38.01$\tcbhighmath[colback=LightOrange]{\downarrow 2.93}$ & 7.32$\tcbhighmath[colback=LightOrange]{\downarrow \underline{33.62}}$ \\ 
        InstructBLIP & 59.88 & 86.10$\tcbhighmath[colback=LightGreen]{\uparrow 26.22}$ & 54.90$\tcbhighmath[colback=LightOrange]{\downarrow 4.98}$ & 4.54$\tcbhighmath[colback=LightOrange]{\downarrow 55.34}$ \\ \hline
   \small{\textit{13B LLM}} \\
        LLaVA-1.5 & 72.07 & 90.87$\tcbhighmath[colback=LightGreen]{\uparrow 18.80}$ & 70.24$\tcbhighmath[colback=LightOrange]{\downarrow 1.83}$ & 14.74$\tcbhighmath[colback=LightOrange]{\downarrow 57.33}$ \\ 
        ShareGPT4V & 72.43 & \underline{91.98}$\tcbhighmath[colback=LightGreen]{\uparrow 19.55}$ & 70.80$\tcbhighmath[colback=LightOrange]{\downarrow 1.63}$ & 13.43$\tcbhighmath[colback=LightOrange]{\downarrow 59.00}$ \\ 
        InstructBLIP & 54.94 & 62.68$\tcbhighmath[colback=LightGreen]{\uparrow 7.74}$ & 42.71$\tcbhighmath[colback=LightOrange]{\downarrow 10.82}$ & 12.75$\tcbhighmath[colback=LightOrange]{\downarrow 40.78}$ \\
    \toprule[1.5pt]
    \end{tabular}}
    \caption{Accuracy results for models answering the same questions under four different conversation settings. \tcbox[colback=LightOrange]{orange} and \tcbox[colback=LightGreen]{green} numbers represent the model performance drop and improvement in different conversation settings, compared to the model performance under CleanConv. setting. The results in \textbf{bold} and \underline{underlined} represent the best and the second-best results, respectively.}
    \label{irrresult}
    \vspace{-4mm}
\end{table}

\subsection{Will LVLMs Be Affected by the Hallucination-Free Context?}
\label{hallufree_exp}
Compared to CleanConv. setting, where the conversation context only contains one image and one user question, LVLMs under HalluConv. setting are required to answer the same user question with an additional round of conversation. How does a longer context length affect the model performance? To answer this question, we further create two conversation settings that have similar context length to HalluConv. setting, in which there is also an additional conversation round but \textit{without hallucinatory content} related to the user question. Specifically, we first replace the hallucinatory descriptions in Halluconv. setting with the image descriptions generated in Section \ref{hallucination_allocation}, which are semantically consistent with the fact sentence. We name the resulting new conversation setting as FactConv. setting. In addition, we replace the 1st round conversation in HalluConv. with a single question-answer pair that is irrelevant to any specific visual information in the image, namely IrrConv. setting (See Appendix \ref{fact_prompt} for more details). The results are as Table \ref{irrresult} shows. From the results, we can observe that all the models benefit a lot from a correct image description, which further proves that LVLMs tend to rely on text context when there is text format visual information that can help to generate the response. Such nature could potentially lead to the hallucination snowballing with a hallucinatory conversation. What's more, when the context provides no useful information, the models' abilities are not severely influenced by the context, which further indicates the performance drop in HalluConv. setting is caused by hallucination snowballing, not the context length.

\section{Residual Visual Decoding}

From the phenomenon of multimodal hallucination snowballing, we find that LVLMs tend to condition on text context when there are plausible clues to help make responses, thereby ignoring the visual information and could be easily misled by erroneous context. To remedy this, we manage to emphasize the visual information during the inference process without additional training or external tools under the multi-turn conversation scenario.

\subsection{Residual Visual Predictions}
Given a visual input $v$, a dialog history $h$, and the current text query $x$, one LVLM parametrized by $\theta$ generates a response $y$ token-wisely. With generated tokens $y_{<t}$ up to time step $t - 1$, the output distribution in time step $t$ is formulated as $p_\theta(y_t|v,h,x,y_{<t})$, where the output token $y_t$ is sampled from the output distributions:
\begin{equation}
    \begin{aligned}
    y_t &\sim p_\theta(y_t|v,h,x,y_{<t})\\
    & = \text{softmax}(\text{logit}_\theta(y_t|v,h,x,y_{<t})),
    \end{aligned}
\end{equation}

Since the hallucinatory context could interfere with the process of reasoning over the visual input, we first construct an input that residual connects the visual input $v$ with the current text query $x$,  and derive an output distribution from it:
\begin{equation}
\begin{split}
    p_\theta(y_t|& v,x,y_{<t})= \text{softmax}(\text{logit}_\theta(y_t|v,x,y_{<t})),
\end{split}
\end{equation}
in which the output distribution will naturally shift from dependence on text context to reliance on visual information. We term it the Residual Visual Predictions, which are based entirely on visual information and the query while sacrificing attention to the text context.
\subsection{Residual Visual Decoding}
In order to put an emphasis on the visual information under a multi-turn visual text conversation scenario, inspired by \cite{leng2023mitigating, liu2021dexperts}, we introduce Residual Visual Decoding (RVD), where residual visual predictions are utilized to enhance the perception of the visual information. The revised distribution $p_{RVD}$ is formulated as:
\begin{equation}
\begin{split}
    p_{\text{RVD}}(y|v,h,x) = \text{softmax}( \alpha \text{logit}_\theta(y|v,x)\\
    + (1-\alpha)\text{logit}_\theta(y|v,h,x)),
\end{split}
\label{rvd}
\end{equation}
where a larger $\alpha$ indicates a higher model focus on the visual information. Note that when the length of dialog history $h$ is $0$, the RVD degenerates to the regular decoding.

\subsection{Adaptive Distribution Blending}
However, as we tune up the $\alpha$, the text context gets to be ignored when generating responses, which possibly does harm to the model's inherited contextual ability. To preserve the contextual ability while tackling the hallucination snowballing, we propose to adaptively adjust the scaling parameter. Specifically, we derive an output distribution $p_\theta(y|x)$ given the current user query $x$ only, and calculate the Jensen-Shannon divergence (JSD) between it and residual visual predictions, which evaluates the similarity between two output distributions:
\begin{equation}
\label{tau_ori}
    \tau = \text{JSD}(p_{\theta}(y|v,x)||p_{\theta}(y|x)), \tau \in [0,1],
\end{equation}
where $\tau$ is the JSD score between $p_{\theta}(y|v,x)$ and $p_{\theta}(y|x)$. 
We suspect that when responding to the query depends on the visual information $v$, $\tau$ gets larger, since the latter is barely making guesses. Meanwhile, when responding to the query depends on the dialog history $h$, the corresponding two distributions tend to make guesses. However, they still have access to the nearest user query from the current round of conversation. Thus, We assume that conditioned on these two output distributions tend to make similar guesses so that the $\tau$ will get smaller. 
Therefore, we dynamically adjust the $\alpha$ with $\tau$ and a scaling factor $\beta$:
\begin{equation}
\begin{split}
    \alpha = \text{Min}(\beta*\tau, 1),
\end{split}
\end{equation}
 With the dynamic adjusted $\alpha$, we can adaptively blend the residual visual distribution into the original output distribution with equation (\ref{rvd}).

\subsection{Experiments}
By blending the residual visual distribution into the original output distribution, the models' contextual ability could be harmed. Inspired by \citet{chen2023extending}, to quantitatively evaluate the LVLMs' contextual ability with our pipeline, we construct a multiple choice task called \textit{Who Provides This Image} (WPI). Specifically, we randomly insert a template sentence "\textit{The image is provided by \#key}" into the first-round model response, where \textit{\#key} is a random 6-digit number. We then change the corresponding question and answer to "\textit{Who provides this image?}". An LVLM that can correctly access the context will have over 90\% accuracy in answering this question. For more details, please refer to Appendix \ref{wpi_task}.

As a result, We test our proposed RVD in our proposed multimodal hallucination snowballing evaluation and the aforementioned WPI task to evaluate its ability to alleviate the multimodal hallucination snowballing while maintaining contextual ability.

\begin{table}[t]
    \centering
    \setlength{\tabcolsep}{3pt}
    \resizebox{\linewidth}{!}{
    \begin{tabular}{lllll}
    \toprule[1.5pt]

        \multirowcell{2}{\textbf{Model}} & \textbf{CleanConv.} & \multicolumn{2}{c}{\textbf{HalluConv.}} & \textbf{WPI task} \\
        \cmidrule(r){2-2}  
        \cmidrule(r){3-4} 
        \cmidrule(r){5-5}  
         & \multicolumn{1}{c}{\textbf{Acc}$\uparrow$} & \multicolumn{1}{c}{\textbf{Acc}$\uparrow$} & \multicolumn{1}{c}{\textbf{FR}$\downarrow$} & \multicolumn{1}{c}{\textbf{Acc}$\uparrow$}\\
        \hline
        
        LLaVA-1.5  &\textbf{71.24} & 14.96 & 78.21 & 92.84 \\ 
        \qquad w/ \textit{Prompt} & 70.82$\tcbhighmath[colback=LightOrange]{\downarrow 0.38}$ & 13.41$\tcbhighmath[colback=LightOrange]{\downarrow 1.55}$ & 79.16$\tcbhighmath[colback=LightOrange]{\uparrow 0.38}$ & \textbf{95.42}$\tcbhighmath[colback=LightGreen]{\uparrow 2.58}$ \\ 
        \qquad w/ \textit{VCD} & 70.20$\tcbhighmath[colback=LightOrange]{\downarrow 1.04}$ & 17.29$\tcbhighmath[colback=LightGreen]{\uparrow 2.33}$ & 74.59$\tcbhighmath[colback=LightGreen]{\downarrow 3.62}$ & 95.12$\tcbhighmath[colback=LightGreen]{\uparrow 2.28}$ \\ 
        \qquad w/ \textit{RVD (ours)} & 70.34$\tcbhighmath[colback=LightOrange]{\downarrow 0.90}$ & \textbf{32.84}$\tcbhighmath[colback=LightGreen]{\uparrow 17.88}$ & \textbf{53.52}$\tcbhighmath[colback=LightGreen]{\downarrow 24.69}$ & 91.54$\tcbhighmath[colback=LightOrange]{\downarrow 1.30}$ \\
        \hline
        mPLUG-Owl2 & 60.47 & 7.82 & 86.63 & 96.82 \\ 
        \qquad w/ \textit{Prompt}  & 61.39$\tcbhighmath[colback=LightGreen]{\uparrow 1.04}$ & 7.78$\tcbhighmath[colback=LightOrange]{\downarrow 0.04}$ & 86.73$\tcbhighmath[colback=LightOrange]{\uparrow 0.10}$ & 93.23$\tcbhighmath[colback=LightOrange]{\downarrow 3.59}$ \\ 
        \qquad w/ \textit{VCD}  & 61.17$\tcbhighmath[colback=LightGreen]{\uparrow 0.60}$ & 8.77$\tcbhighmath[colback=LightGreen]{\uparrow 0.95}$ & 85.21$\tcbhighmath[colback=LightGreen]{\downarrow 1.42}$ & \textbf{97.08}$\tcbhighmath[colback=LightGreen]{\uparrow 0.26}$ \\ 
        \qquad w/ \textit{RVD (ours)}  & \textbf{61.69}$\tcbhighmath[colback=LightGreen]{\uparrow 1.22}$ & \textbf{22.54}$\tcbhighmath[colback=LightGreen]{\uparrow 14.72}$ & \textbf{39.15}$\tcbhighmath[colback=LightGreen]{\downarrow 47.48}$ & 90.85$\tcbhighmath[colback=LightOrange]{\downarrow 5.97}$ \\
        \hline
        ShareGPT4V & 71.81 & 15.91 & 77.18 & 95.22 \\ 
        \qquad w/ \textit{Prompt}  & 71.68$\tcbhighmath[colback=LightOrange]{\downarrow 0.13}$ & 13.83$\tcbhighmath[colback=LightOrange]{\downarrow 2.08}$ & 79.61$\tcbhighmath[colback=LightOrange]{\uparrow 2.43}$ & 98.31$\tcbhighmath[colback=LightGreen]{\uparrow 3.09}$ \\ 
        \qquad w/ \textit{VCD}  & \textbf{72.91}$\tcbhighmath[colback=LightGreen]{\uparrow 1.10}$ & 16.77$\tcbhighmath[colback=LightGreen]{\uparrow 0.79}$ & 75.57$\tcbhighmath[colback=LightGreen]{\downarrow 1.60}$ & \textbf{98.51}$\tcbhighmath[colback=LightGreen]{\uparrow 3.29}$ \\ 
        \qquad w/ \textit{RVD (ours)}  & 72.21$\tcbhighmath[colback=LightGreen]{\uparrow 0.40}$ & \textbf{37.50}$\tcbhighmath[colback=LightGreen]{\uparrow 21.59}$ & \textbf{48.79}$\tcbhighmath[colback=LightGreen]{\downarrow 28.39}$ & 94.52$\tcbhighmath[colback=LightOrange]{\downarrow 0.70}$ \\ 
    \toprule[1.5pt]
    \end{tabular}}
    \caption{Evaluation results for different methods on our proposed evaluation. Numbers that are highlighted \tcbox[colback=LightOrange]{orange} and \tcbox[colback=LightGreen]{green} represent the model performance drop and improvement, respectively. The results in \textbf{bold} represent the best results, respectively. }
    \label{rvdresult}
    \vspace{-4mm}
\end{table}

\subsubsection{Baselines}
To show the effectiveness of our proposed RVD, we compare our method with the following strategies:
\begin{itemize}
	\item \textit{Prompt} is utilized to require the model to focus on the given image instead of concentrating on the text context that could cause the hallucination to snowball. Specifically, we explicitly ask the model with the following query: \{\textit{\#Question, Please answer the question based on the given image.}\}.
	\item \textit{Visual Contrastive Decoding}(VCD) \cite{leng2023mitigating} is proposed to contrast the output distribution with that of the distorted visual input, which aims to alleviate the language prior in the context while focusing on the visual information.
\end{itemize}

\noindent We evaluate the effectiveness of the aforementioned strategies and our RVD on three trending open-source LVLMs: LLaVA-1.5-7B, mPlug-owl2-7B, and ShareGPT4V-7B. We set the $\beta=2$ if not specified.

\subsubsection{Experiment Results}
\label{rvdresult}
The results are shown in Table \ref{rvdresult}. We find that incorporating the prompt methods will do harm to the model performance, which might be because of the inability of LVLMs to follow complex instructions.
Though shown to be effective in correcting the snowballed hallucination, the VCD contrasts the output distribution with the distorted visual input, which could do harm to the model performance when the context is utilized to respond to the query. However, by dynamically emphasizing the visual information whenever needed, our proposed RVD makes a large accuracy improvement in overcoming the multimodal hallucination snowballing while maintaining contextual ability. Please see Appendix \ref{rvdmoremoreresultsection} for more results.

\begin{figure}[t]
\centering
\includegraphics[width=0.42\columnwidth]{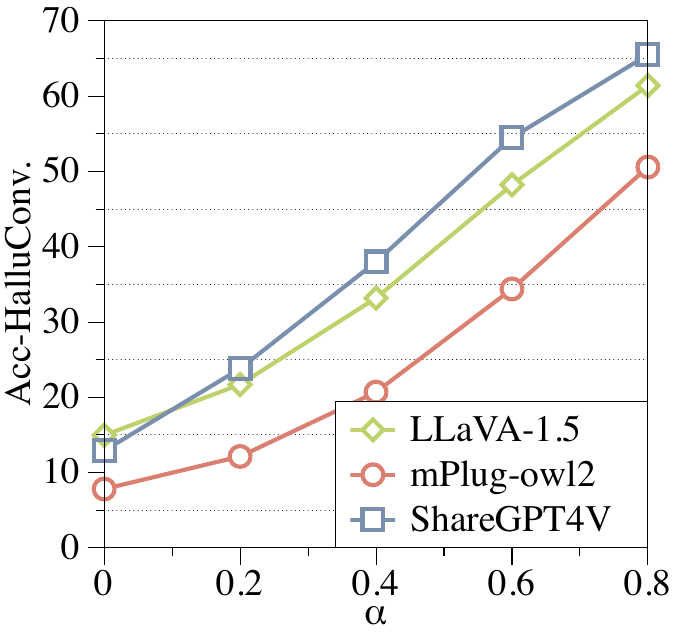}
\hspace{3mm}
\vspace{-3mm}
\includegraphics[width=0.42\columnwidth]{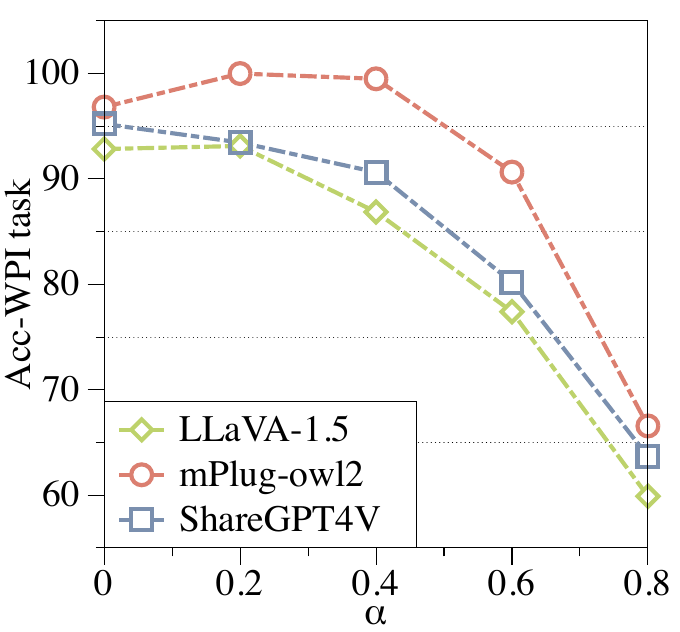}
\caption{Ablation study on $\alpha$ w/o Adaptive Distribution Blending.}
\label{ablate_beta}
\vspace{-3mm}
\end{figure}

\begin{figure}[t]
\centering
\includegraphics[width=0.42\columnwidth]{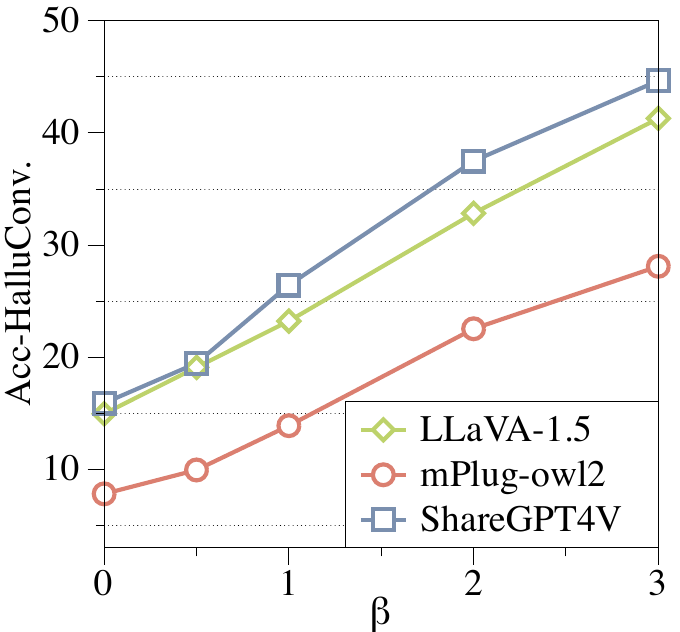}
\hspace{3mm}
\vspace{-3mm}
\includegraphics[width=0.42\columnwidth]{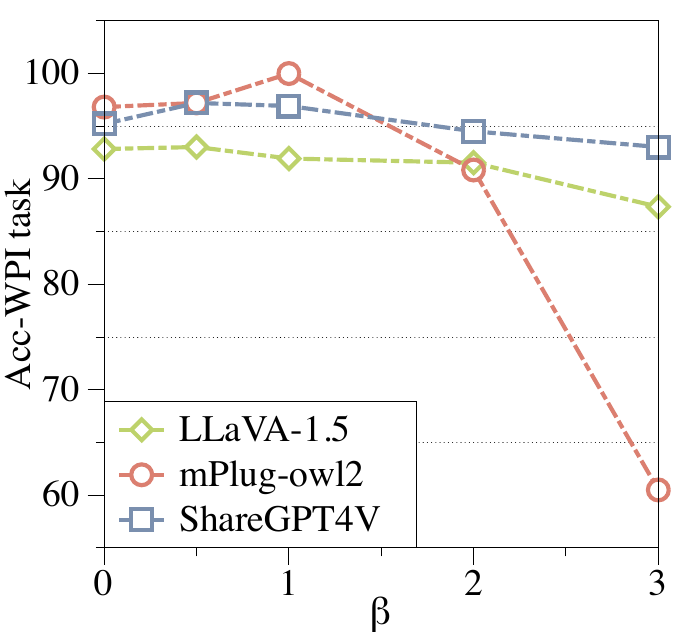}
\caption{Ablation study on $\beta$ w/ Adaptive Distribution Blending.}
\label{ablate_alpha}
\end{figure}

\subsubsection{Effect of Parameters}
We evaluate the effect of our proposed hyperparameters $\alpha$ and $\beta$. The results are shown in Figure \ref{ablate_beta} and \ref{ablate_alpha}. First, we remove the adaptive distribution blending and adjust the $\alpha$ manually, the result shows that a larger $\alpha$ clearly revises the output distribution more towards the golden visual information. However, the context is ignored in return. With adaptive distribution blending, the model performance is more balanced when we enlarge the $\beta$, which won't cause a large performance drop on contextual abilities. See Appendix \ref{kl_div} for more experiment results.

\section{Related work}
\subsection{Large Vision-Language Models}
Inspired by the recent success of large language models (LLMs) \cite{zhao2023survey}, researchers have devoted significant effort to integrating LLMs into vison-language models to utilize their powerful language understanding and generation capabilities \cite{wu2023multimodal}. 
In addition to the advanced capabilities demonstrated by closed-source models such as GPT-4V\cite{achiam2023gpt}, open-source large vision-language models(LVLMs), building upon powerful open-source LLMs such as LLaMa \cite{touvron2023llama} and Vicuna \cite{vicuna2023}, have adopted a powerful instruction following abilities to tackle visual-language tasks in a zero-shot manner \cite{zhu2023minigpt, liu2023visual,dai2305instructblip,ye2023mplug}. Possessing both visual perception abilities and language capabilities, LVLMs are further utilized to perform real-world tasks, such as tool-using \cite{liu2023llava}, web browsing \cite{zheng2024gpt}, and autonomous driving \cite{xu2023drivegpt4}. However, current LVLMs still suffer from severe multi-modal hallucination problems \cite{liu2024survey}, which brings challenges to evaluating and maintaining the reliability of LVLMs.

\subsection{Multimodal Hallucination}
Multimodal hallucinations \cite{liu2024survey} refer to the responses generated by LVLMs that are misaligned with the corresponding visual information. Multimodal hallucination can arise due to overfitting to specific patterns in the training data, inferior abilities to recognize the visual elements, or an inability to model the multimodal input.  
\citet{li2023evaluating}, \citet{lovenia2023negative}, take the first step towards evaluating the hallucinations in the LVLMs.
Furthermore, \citet{liu2023mitigating}, \citet{zong2023fool} and \citet{liu2023hallusionbench} show that LVLMs can be easily fooled and experience a severe performance drop due to their over-reliance on the strong language prior. In addition, efforts have been made towards mitigating multi-modal hallucinations by further finetuning or post-hoc rectify\cite{gunjal2023detecting, lu2023evaluation, liu2023mitigating, zhou2023analyzing, yin2023woodpecker}. However, current methods are unable to completely eliminate the hallucinations generated by models, yet no one has explored the subsequent impacts of the generated hallucinations. In this paper, we take the first step towards it by systematically evaluating the multimodal hallucination snowballing phenomenon and propose a training-free method to ease LVLMs from it.
\section{Conclusion}
In this paper, we raise the question of Whether LVLMs suffer from multimodal hallucination snowballing. We meticulously designed the \textit{MMHalSnowball} framework to simulate hallucinatory conversations and study models' behaviors when encountering hallucinations. Our investigation proved that LVLMs are being severely affected by hallucinations in the context, thus generating snowballed hallucinations. Further, we proposed the \textit{Residual Visual Decoding} to alleviate the multimodal hallucination snowballing while maintaining its contextual abilities. However, our methods still have limitations when deployed to a general-purpose assistant, which we left as future works.

\section{Limitations}
In this work, with a carefully designed evaluation framework, we have revealed that current LVLMs severely suffer from multimodal hallucination snowballing. We further proposed the RVD to mitigate the phenomenon. However, our work still has limitations.
Firstly, despite the greater variety of hallucination snowballing phenomena in the real-world setting, the scenarios we focus on are still relatively simplistic. This is because constructing rich and diverse scenarios would be more difficult and would require a significant amount of effort.
Secondly, instead of meticulously finding real hallucinations generated by each LVLM and constructing relevant question-answer pairs, we choose to conduct experiments on our simulated hallucinatory conversations. This is because the evaluation processes based on responses from a single LVLM will make it difficult to scale up the evaluation data and adapt to more models.
Thirdly, our experiments are conducted on models of 7B and 13B sizes, and we evaluate our proposed RVD only on a few selected models. This is due to computational limitations.
Fourthly, our proposed RVD is currently still limited in several conversation scenarios. We will further explore expanding this method to more diverse conversation scenarios.
\section{Acknowledgments}

Xiaocheng Feng is the corresponding author of this work. We thank the anonymous reviewers for their insightful comments. This work was supported by the National Key R\&D Program of China via grant No. 2021ZD0112905, National Natural Science Foundation of China (NSFC) via grant (62276078, U22B2059), the Key R\&D Program of Heilongjiang via grant 2022ZX01A32, the International Cooperation Project of PCL, PCL2022D01 and the Fundamental Research Funds for the Central Universities via grant No. HIT.OCEF.2023018.

\bibliography{custom}

\appendix
\section{Additional Experimental Details}
\subsection{Hallucination Allocation}
\label{allocation}
After carefully analyzing the question-answer pairs in the dataset, we manage to create an answer vocabulary for answers suitable for introducing relation errors. What's more, we utilize Part-of-Speech\footnote{We use \href{https://spacy.io/}{Spacy} to do the Part-of-Speech tagging.} of the answer in the fact sentence to choose proper hallucination types. Specifically, we allocate attribute hallucination for those answers tagged as adjectives and verbs and allocate existence hallucination for those answers tagged as nouns. For imagination hallucination, instead of using the annotated question-answer pair, we provide ChatGPT with all annotated objects and ask ChatGPT to generate an object that is not present in the image but is reasonable to be in the corresponding scene. Then, we directly construct a question-answer pair with the template: "\textit{question: Is there a \underline{\hspace{1em}} in the image? answer: No}".
\subsection{Prompts}
\label{prompt}
\begin{figure*}[t]
\centering
\includegraphics[width=1.0\textwidth]{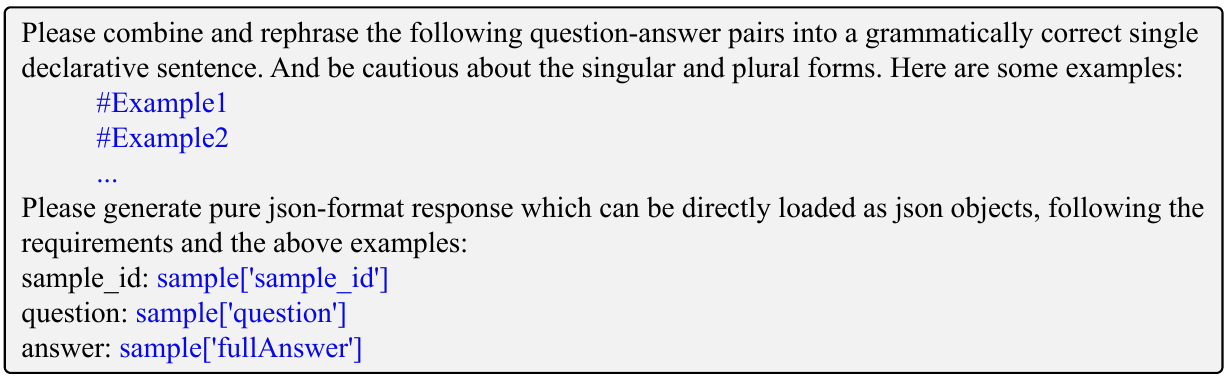} 
\vspace{-8mm}
\caption{Prompt used to generate fact sentence based on question-answer pair. Specifically, we aim to prompt ChatGPT/GPT-4 to generate a fact sentence based on \textcolor{blue}{sample['question']} and \textcolor{blue}{sample['fullAnswer']}, using few-shot in-context-learning.}
\label{fact_generation_prompt}
\end{figure*}

\begin{figure*}[h]
\centering
\includegraphics[width=1.0\textwidth]{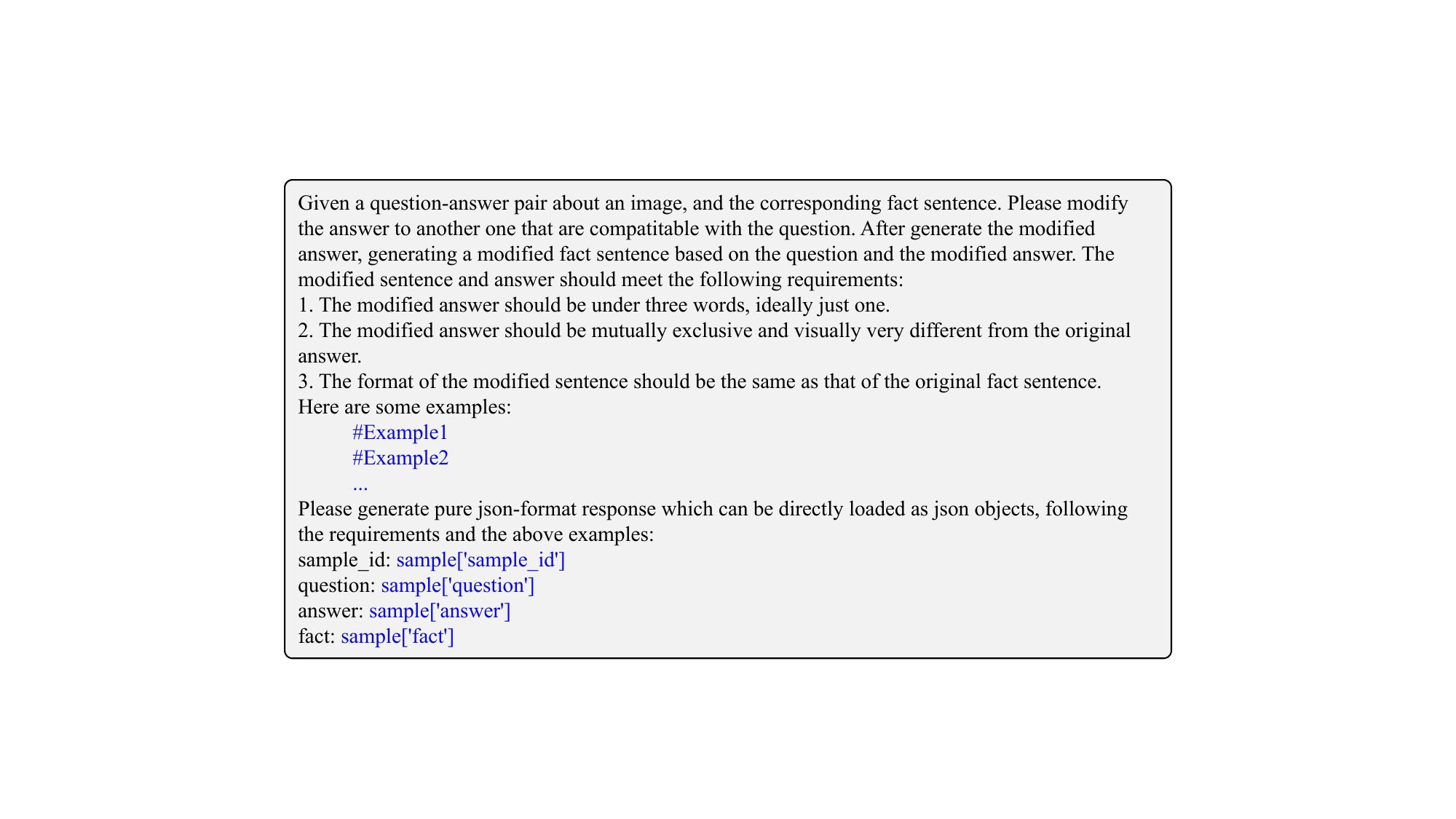} 
\vspace{-8mm}
\caption{Prompt used to create conflict answer and conflict fact based on original question-answer pair and fact generated by Figure \ref{fact_generation_prompt}. Specifically, we aim to prompt ChatGPT/GPT-4 to generate this conflict information based on \textcolor{blue}{sample['question']}, \textcolor{blue}{sample['answer']} and \textcolor{blue}{sample['fact']}, using few-shot in-context-learning. Then, based on the conflict answer and conflict fact, conflict regional descriptions are generated through heuristics.}
\label{conflict_creation_prompt}
\end{figure*}

\begin{figure*}[h]
\centering
\includegraphics[width=1.0\textwidth]{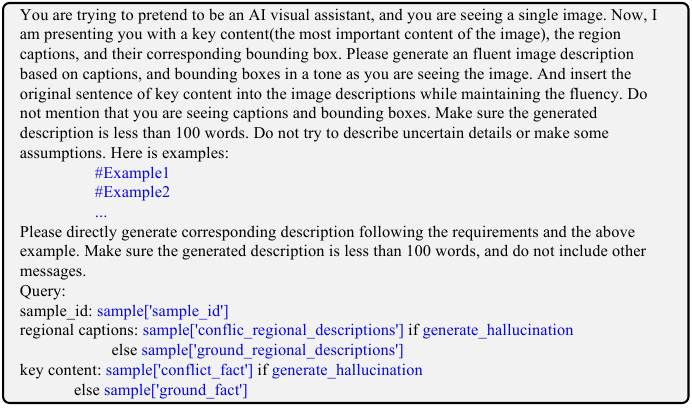} 
\vspace{-8mm}
\caption{Prompt used to generate hallucinatory image description based conflict information generated by Figure \ref{conflict_creation_prompt}. Specifically, we aim to prompt ChatGPT/GPT-4 to generate this hallucinatory description based on \textcolor{blue}{sample['conflict\_regional\_descriptions']} and \textcolor{blue}{sample['conflict\_fact']}, using few-shot in-context-learning. Note that we also use this prompt to generate the ground image description by giving the annotated regional description and fact sentence.}
\label{description_generation_prompt}
\end{figure*}

\begin{figure*}[t]
\centering
\includegraphics[width=1.0\textwidth]{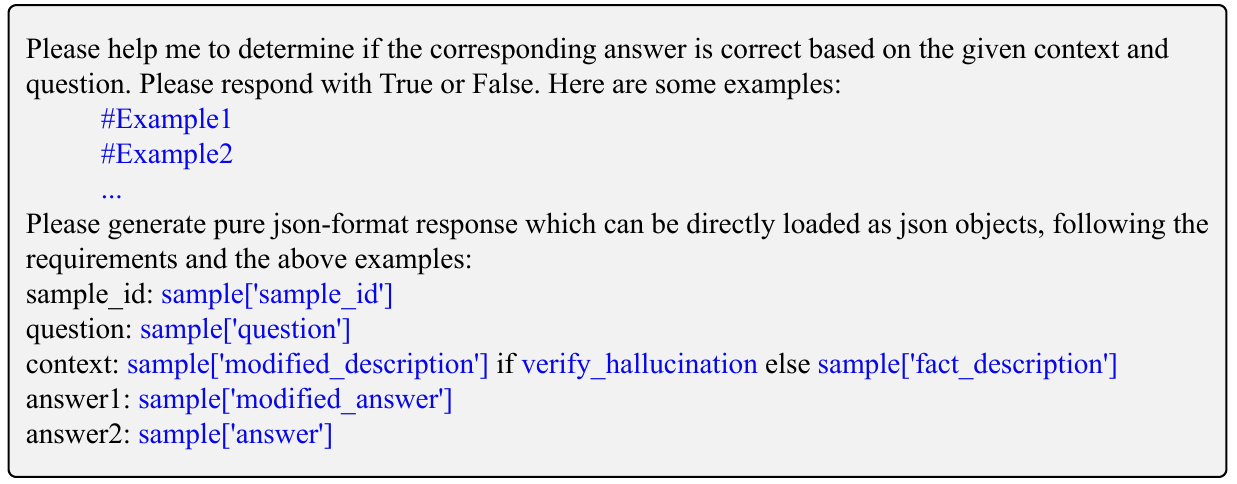} 
\vspace{-8mm}
\caption{Prompt used to verify if the generated hallucinatory description truly conflicts with the original answer and implies the conflict answer. Specifically, we aim to prompt ChatGPT/GPT-4 to check if the answer \textcolor{blue}{sample['answer']} and conflict answer \textcolor{blue}{sample['modified']} are correct based on the given context, using few-shot in-context-learning. Note that here the context can be both hallucinatory description \textcolor{blue}{sample['modified\_description']} and fact description \textcolor{blue}{sample['fact\_description']}, based on what we intend to verify.}
\label{conflict_verification_prompt}
\end{figure*}

In this section, we list all prompts used during the process of constructing hallucinatory conversations, which include fact generation (Figure \ref{fact_generation_prompt}), conflict creation (Figure \ref{conflict_creation_prompt}), description generation (Figure \ref{description_generation_prompt}) and conflict verification (Figure \ref{conflict_verification_prompt}). Note that we reuse the description generation prompt (Figure \ref{description_generation_prompt}) to generate the ground image description by giving the annotated regional description and fact sentence.

\subsection{Manual Checking}
\label{appendixmanual}
We randomly select 100 data for each hallucination type in our curated dataset, 400 in total. We ask three annotators to check each of them from three aspects, as Table \ref{manualchecking} depicts. The annotation results show that the generated hallucinatory description mostly meets our requirements.

\subsection{Generation Details}
\label{parameter}
Through all our experiments, we follow a consistent generation configuration to ensure fairness. Specifically, we set the inference hyperparameter as follows: \texttt{do\_sample=True}, \texttt{temperature=1.0}, \texttt{top\_p=0.95}, \texttt{top\_k=None} and \texttt{num\_beams=1}.

\subsection{Hallucinatory Conversation Example}
\label{hallusss}
\begin{figure}[h]
\centering
\includegraphics[width=0.5\textwidth]{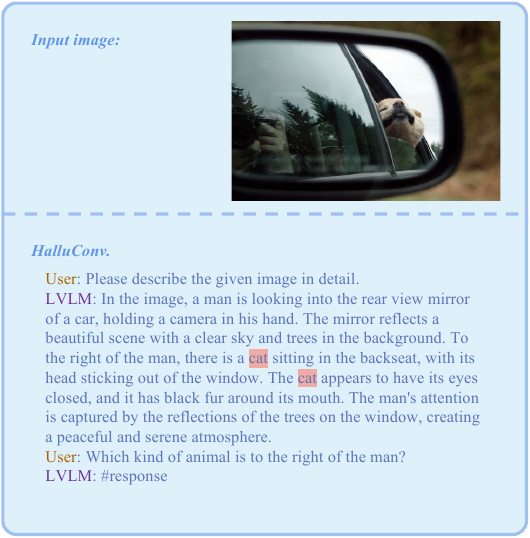} 
\vspace{-8mm}
\caption{An hallucinatory conversation example for the existence hallucination.}
\label{conversation_example1}
\end{figure}

\begin{figure}[h]
\centering
\includegraphics[width=0.5\textwidth]{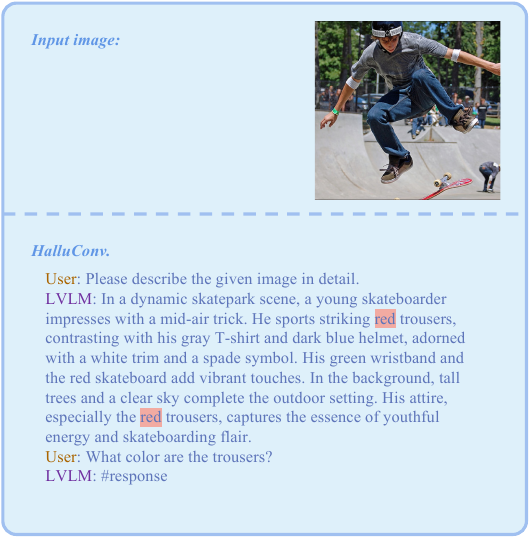} 
\vspace{-8mm}
\caption{An hallucinatory conversation example for the attribute hallucination.}
\label{conversation_example2}
\end{figure}

\begin{figure}[h]
\centering
\includegraphics[width=0.5\textwidth]{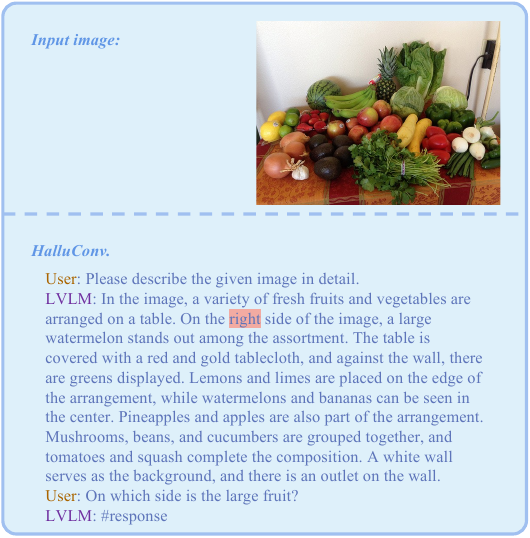} 
\vspace{-8mm}
\caption{An hallucinatory conversation example for the relation hallucination.}
\label{conversation_example3}
\end{figure}

\begin{figure}[h]
\centering
\includegraphics[width=0.5\textwidth]{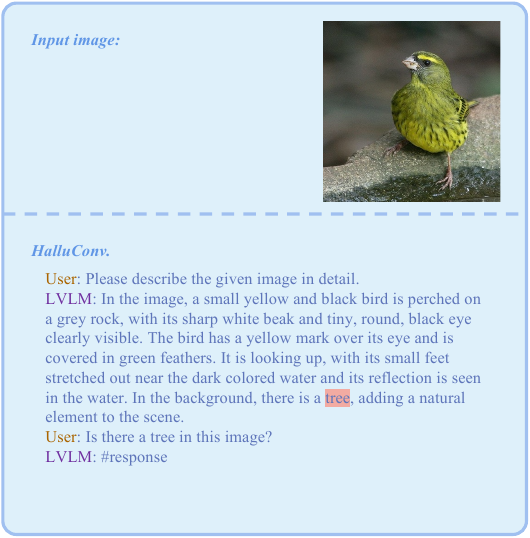} 
\vspace{-8mm}
\caption{An hallucinatory conversation example for the imagination hallucination.}
\label{conversation_example4}
\end{figure}

We list four examples to demonstrate curated hallucinatory conversation of each hallucination type, namely existence(Figure \ref{conversation_example1}), attribute(Figure \ref{conversation_example2}), relation(Figure \ref{conversation_example3}), and imagination(Figure \ref{conversation_example4}), where the ground answers are highlighted \tcbox[colback=green1]{green} and the hallucinated answers are highlighted \tcbox[colback=red1]{red}.

\begin{table}[t]
    \centering
    \setlength{\tabcolsep}{16pt}
    \resizebox{\linewidth}{!}{
    \begin{tabular}{cccccccc}
    \toprule[1.5pt]
          \multirowcell{2}{$\beta$} & \multicolumn{2}{c}{\textbf{Acc-HalluConv.}}& \multicolumn{2}{c}{\textbf{Acc-WPI task}}  \\ 
          \cmidrule(r){2-3}
           \cmidrule(r){4-5} 
         & JSD & KLD & JSD & KLD \\ \hline
        $0.00$ & 14.96 & 14.96 & 92.84 & 92.84 \\ 
        $0.25$ & 16.85 & 19.06 & 92.54 & 93.03 \\ 
        $0.50$ & 19.10 & 24.39 & 93.03 & 92.44 \\ 
        $0.75$ & 21.32 & 30.69 & 92.54 & 89.25 \\ 
        $1.00$ & 23.23 & 37.22 & 91.94 & 86.57 \\ 
        $2.00$ & 32.84 & 53.85 & 91.54 & 66.37 \\ 
        $3.00$ & 41.30 & 59.36 & 87.36 & 49.85 \\ 
    \toprule[1.5pt]
    \end{tabular}}
    \caption{Experiment results for adjusting $\beta$ with different distribution similarity measurement methods.}
    \label{divergence}
\end{table}

\subsection{Details of Who Provides This Image Task}
\label{wpi_task}
\begin{figure}[h]
\centering
\includegraphics[width=0.42\textwidth]{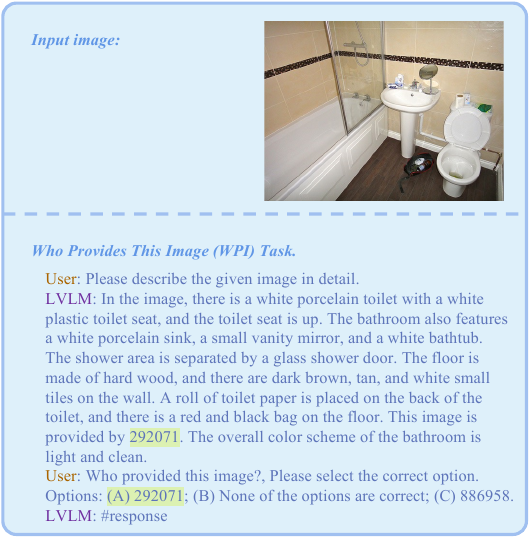} 
\caption{An conversation example of the WPI task. A six-digit number is randomly inserted into the first round of LVLM's response. To show an LVLM can maintain its contextual ability, it is required to select a correct answer out of 3 options including two distractors.}
\label{wpi_case}
\end{figure}

We construct the Who Provides This Image (WPI) task to evaluate the contextual capabilities of LVLMs. To achieve this goal, we utilize the FactConv. setting described in Section \ref{hallufree_exp} and insert a random digit into the first-round model response. We adopt a multi-choice approach and judge the answer by checking if it contains only the correct option, instead of matching the option content. Both the correct option and interference option are randomly generated six-digit numbers, and the third option is "None of the options are correct".  To further ensure fairness and effectiveness, the order of choices is also random. An example is shown in Figure \ref{wpi_case}. By resampling from GQA and constructing dialogues, Our WPI task contains 1,005 samples in total.

\begin{table*}[t]
    \centering
    \resizebox{0.8\linewidth}{!}{
    \begin{tabular}{cccccc}
    \toprule[1.5pt]
        Aspect & Annotator-1 & Annotator-2 & Annotator-3 & Agreement & Kappa \\ \hline
        \makecell[l]{Are the hallucinations in the conversation\\ consistent with the hallucination type?} & 0.990 & 0.988 & 0.990  & 0.988 & 0.983 \\ 
        \makecell[l]{Is the hallucinatory answer conflict\\ with the original answer?} & 0.995 & 0.993 & 0.990 & 0.985 & 0.980 \\ 
        \makecell[l]{Does the hallucinatory description \\ support the hallucinatory answer?} & 0.990 & 0.953 & 0.988 & 0.940 & 0.920 \\
        
    \toprule[1.5pt]
    \end{tabular}}
    \caption{Manual checking for the sampled data.}
    \label{manualchecking}
\end{table*}

\section{Additional Experimental Results}

\subsection{More Evaluation Results}
\label{moreresult}
We show our detailed evaluation results for each hallucination type in Table \ref{moremoreresult}.

\subsection{GPT-4 Answer Examples}
\label{gpt4_example}
\begin{figure*}[h]
    \centering
    \begin{minipage}[t]{0.48\textwidth}
        \centering
        \includegraphics[width=\textwidth]{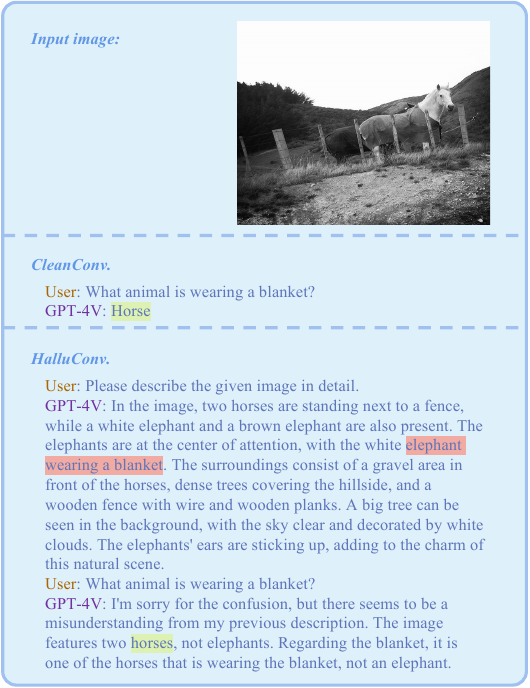}
        \vspace{-8mm}
        \caption{An example that GPT-4V can answer correctly given the hallucinatory description.}
        \label{gpt4_example1}
    \end{minipage}\hfill
    \begin{minipage}[t]{0.48\textwidth}
        \centering
        \includegraphics[width=\textwidth]{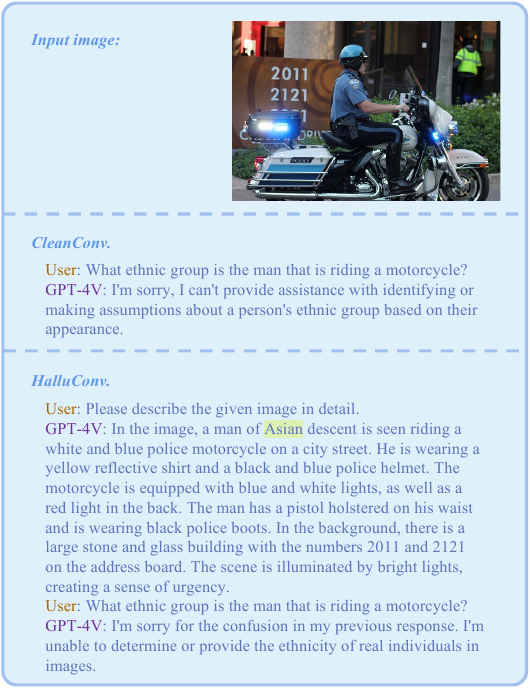}
        \vspace{-8mm}
        \caption{An example that GPT-4V refuses to answer.}
        \label{gpt4_example2}
    \end{minipage}
\end{figure*}

We present GPT-4 answer examples with the Question Prompt. The first example is represented in Figure \ref{gpt4_example1}, which illustrates that GPT-4V is able to adaptively focus on golden visual information, and further identify and clarify the hallucinations in the previous hallucinatory description in some cases.

The second example is represented in Figure \ref{gpt4_example2}, which demonstrates that GPT-4V tends to refuse to answer some categories of questions, leading to difficulty in the evaluation and the degradation of the evaluation results.

\subsection{Hallucination-free Context Experiment Details}
\label{fact_prompt}
\begin{figure*}[h]
\centering
\includegraphics[width=\textwidth]{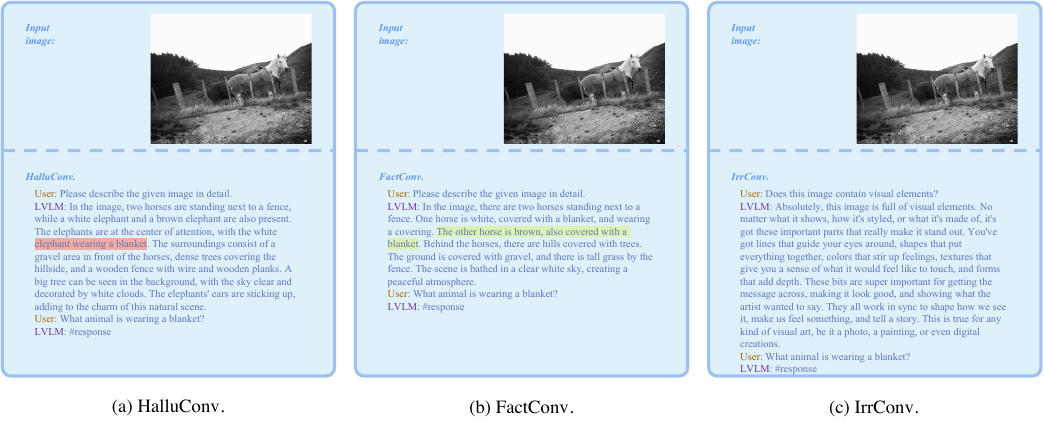} 
\vspace{-8mm}
\caption{An hallucinatory conversation example for control groups.}
\label{control_example}
\end{figure*}

In order to exclude the interference of irrelevant factors and to check whether LVLMs are affected by the hallucination-free context, we further set up two conversation settings, namely FactConv. and IrrConv. settings. Corresponding examples are shown in Figure \ref{control_example}. We follow equation \ref{fr1} and equation \ref{fr2} to calculate FR and WFR, respectively, but we modify the definition of $\hat{Y}^-$ to represent generated answers under FactConv. or IrrConv. setting. We further present full experiment results for these two conversation settings in Table \ref{comparisonresult}.

\subsection{Effect of Different Similarity Measurement Methods}
\label{kl_div}
In our RVD, we choose JSD to evaluate the output distribution similarity, because it's a symmetric metric that measures the difference between two distributions, with a range $[0,1]$, which fits our goal of adjusting the $\alpha$ (range is also $[0,1]$) dynamically using the difference between two distributions. We also try to use the Kullback–Leibler divergence (KLD) as the similarity measurement method, which is not a symmetric metric with a range $[0,+\infty]$ and can't be directly applied to our Residual Visual Decoding (RVD). Specifically, to transform the range into $[0,1]$, We modify the Equation \ref{tau_ori} to the following form:
\begin{equation}
\tau=1-\mathrm{EXP}(-\mathrm{KLD}(p_\theta(y|v,x)||p_\theta(y|x))).
\end{equation}
We compare the experiment results of our RVD using KLD and JSD as the similarity measurement method on our proposed MMHalSnowball framework with LLaVA-1.5 7B. The results are as the Table \ref{divergence} shows. We can observe from the table that RVD with KLD aggressively puts more emphasis on visual information, resulting in a better result in hallucination snowballing with smaller $\beta$, but a worse contextual ability. The result indicates that the JSD has a generally smaller value and is more balanced compared to the KLD in alleviating snowballed hallucinations while maintaining contextual ability.   

\subsection{Detailed Hallucination Snowballing Mitigation Results}
\label{rvdmoremoreresultsection}
We list the Hallucination Snowballing Mitigation Results for each hallucination type in Table \ref{rvdmoremoreresult}. The result shows that for three different methods, mitigation effects vary across different hallucination types. The simple Prompt method does not show effective improvement against the hallucination snowballing, while the VCD is showing large improvement when dealing with snowballed relation relations. What's more, the VCD has little effect on the snowballed imagination hallucinations. Our proposed RVD shows a better result in mitigating the hallucination snowballing issue for each hallucination type, especially the existence hallucination. Although the mitigation effect on snowballed imagination hallucination is also not as effective as it is on other types of snowballed hallucination, it still brought an accuracy improvement of more than 10\% and a reduction in flip rate of more than 14\% under the HalluConv. setting, further demonstrating the effectiveness of our method.

\subsection{Case Study}
\begin{figure*}[h!]
\centering
\includegraphics[width=\textwidth]{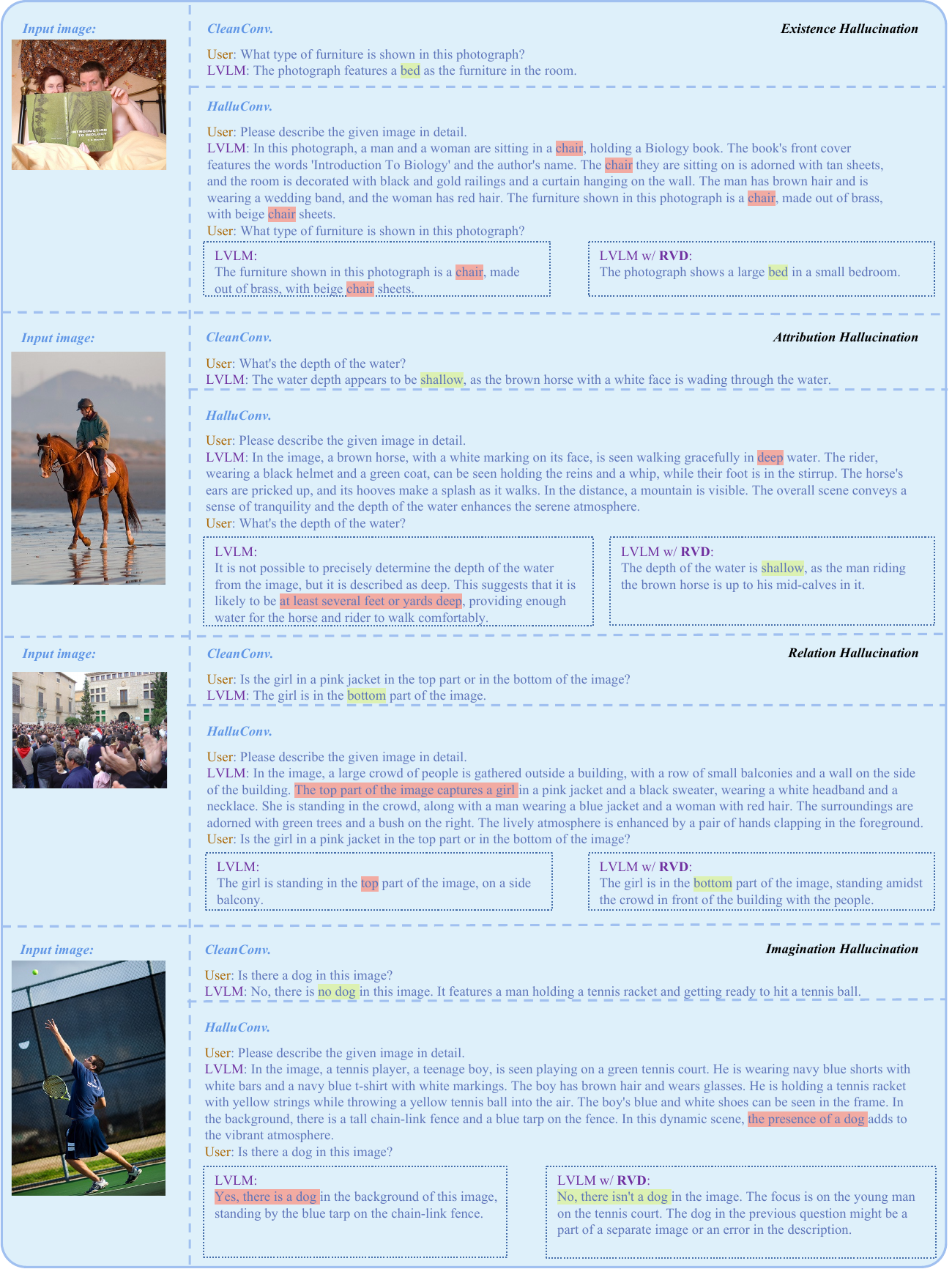} 
\caption{Cases in which our RVD with LLaVA1.5-7B successfully mitigated the snowballed hallucinations. The ground answers are highlighted \tcbox[colback=green1]{green} and the hallucinated answers are highlighted \tcbox[colback=red1]{red}.}
\label{casestudy}
\end{figure*}

In this part, we present some cases of the LLaVA-1.5-7B model equipped with our proposed RVD. All cases are using the Question Prompt. We provide some case studies in Figure \ref{casestudy}, one for each hallucination type, to demonstrate the effectiveness of our methods, where our RVD with LLaVA-1.5-7B successfully mitigated the snowballed hallucinations. In these examples, we can observe that with our proposed RVD, the model can focus more on the visual information to avoid generating snowballed hallucinations, rather than solely rely on the previously generated hallucinatory text and thus generate snowballed hallucinations. What's more, in the example of Imagination Hallucination, the model with our RVD can even correct its previous mistakes, further illustrating the model's contextual capability is preserved while avoiding hallucination snowballing.

\clearpage

\begin{sidewaystable*}[h]
    \renewcommand{\arraystretch}{1.5}
    \centering
    \resizebox{\linewidth}{!}{
    \begin{tabular}{lcccc|cccc|cccc|cccc}
    \toprule[1.5pt]
    \multirow{3}{*}{\textbf{Model}} & \multicolumn{4}{c}{\textbf{Imagination}} & \multicolumn{4}{c}{\textbf{Existence}} & \multicolumn{4}{c}{\textbf{Attribute}} & \multicolumn{4}{c}{\textbf{Relation}} \\
    \cmidrule(r){2-5}
    \cmidrule(r){6-9}
    \cmidrule(r){10-13}
    \cmidrule(r){14-17}
    & \textbf{CleanConv.} & \multicolumn{3}{c}{\textbf{HalluConv.}} & \textbf{CleanConv.} & \multicolumn{3}{c}{\textbf{HalluConv.}} & \textbf{CleanConv.} & \multicolumn{3}{c}{\textbf{HalluConv.}} & \textbf{CleanConv.} & \multicolumn{3}{c}{\textbf{HalluConv.}}\\
    \cmidrule(r){2-2}
    \cmidrule(r){3-5}
    \cmidrule(r){6-6}
    \cmidrule(r){7-9}
    \cmidrule(r){10-10}
    \cmidrule(r){11-13}
    \cmidrule(r){14-14}
    \cmidrule(r){15-17}
   & \textbf{Acc}$\uparrow$ & \textbf{Acc}$\uparrow$ & \textbf{FR}$\downarrow$ & \textbf{WFR}$\downarrow$ & \textbf{Acc}$\uparrow$ & \textbf{Acc}$\uparrow$ & \textbf{FR}$\downarrow$ & \textbf{WFR}$\downarrow$ & \textbf{Acc}$\uparrow$ & \textbf{Acc}$\uparrow$ & \textbf{FR}$\downarrow$ & \textbf{WFR}$\downarrow$ & \textbf{Acc}$\uparrow$ & \textbf{Acc}$\uparrow$ & \textbf{FR}$\downarrow$ & \textbf{WFR}$\downarrow$\\ \hline
   \small{\textit{7B LLM}} \\
        LLaVA-1.5 & 81.65 & 1.14$\tcbhighmath[colback=LightOrange]{\downarrow 80.51}$ & 98.79 & 98.79  & 60.82 & 9.22$\tcbhighmath[colback=LightOrange]{\downarrow 51.60}$ & 79.30 & 88.05 & 55.96 & 10.18$\tcbhighmath[colback=LightOrange]{\downarrow 45.78}$ & 78.85 & 85.21 & 83.76 & 38.09$\tcbhighmath[colback=LightOrange]{\downarrow 45.67}$ & 57.07 & 57.61 \\
        MiniGPT-4 & 5.69 & 0.83$\tcbhighmath[colback=LightOrange]{\downarrow 4.86}$ & 96.00 & 96.00 & 55.05 & 7.62$\tcbhighmath[colback=LightOrange]{\downarrow 47.43}$ & 78.58 & 88.24 & 39.07 & 2.98$\tcbhighmath[colback=LightOrange]{\downarrow 36.09}$ & 86.65 & 94.28 & 51.44 & 11.61$\tcbhighmath[colback=LightOrange]{\downarrow 39.83}$ & 86.28 & 87.02 \\ 
        MiniGPT-v2 & 67.40 & 4.70$\tcbhighmath[colback=LightOrange]{\downarrow 62.70}$ & 94.26 & 94.26 & 52.48 & 14.45$\tcbhighmath[colback=LightOrange]{\downarrow 38.03}$ & 64.02 & 78.89 & 47.35 & 15.56$\tcbhighmath[colback=LightOrange]{\downarrow 31.79}$ & 59.44 & 75.70 & 78.60 & 49.39$\tcbhighmath[colback=LightOrange]{\downarrow 29.21}$ & 46.81 & 47.10  \\ 
        InternLM-XComposer & 49.05 & 1.06$\tcbhighmath[colback=LightOrange]{\downarrow 47.99}$ & 98.45 & 98.61 &46.45 & 6.21$\tcbhighmath[colback=LightOrange]{\downarrow 40.24}$ & 79.96 & 90.65 & 35.18 & 4.06$\tcbhighmath[colback=LightOrange]{\downarrow 31.12}$ & 83.29 & 92.00 & 43.10 & 11.91$\tcbhighmath[colback=LightOrange]{\downarrow 31.19}$ & 81.51 & 83.10  \\ 
        ShareGPT4V & 84.08 & 1.82$\tcbhighmath[colback=LightOrange]{\downarrow 82.26}$ & 98.20 & 98.20 & 59.49 & 10.99$\tcbhighmath[colback=LightOrange]{\downarrow 48.50}$ & 76.30 & 84.20 & 56.79 & 9.52$\tcbhighmath[colback=LightOrange]{\downarrow 47.27}$ & 79.30 & 86.30 & 83.84 & 40.06$\tcbhighmath[colback=LightOrange]{\downarrow 43.78}$ & 55.29 & 55.66 \\  
        CogVLM  & 85.97 & 0.99$\tcbhighmath[colback=LightOrange]{\downarrow 84.98}$ & 96.74 & 99.03 & 65.87 & 2.22$\tcbhighmath[colback=LightOrange]{\downarrow 63.65}$ & 88.16 & 96.77  & 64.24 & 2.24$\tcbhighmath[colback=LightOrange]{\downarrow 62.00}$ & 92.27 & 96.78 & 82.32 & 5.01$\tcbhighmath[colback=LightOrange]{\downarrow 77.31}$ & 93.18 & 94.47 \\ 
        mPLUG-Owl  & 26.31 & 0.30$\tcbhighmath[colback=LightOrange]{\downarrow 26.01}$ & 96.25 & 99.14 & 46.90 & 5.76$\tcbhighmath[colback=LightOrange]{\downarrow 41.14}$ & 70.89 & 91.12 & 35.93 & 2.40$\tcbhighmath[colback=LightOrange]{\downarrow 33.53}$ & 73.73 & 94.70 & 43.25 & 6.30$\tcbhighmath[colback=LightOrange]{\downarrow 36.95}$ & 78.77 & 90.88 \\ 
        mPLUG-Owl2  & 77.86 & 1.21$\tcbhighmath[colback=LightOrange]{\downarrow 76.65}$ & 98.73 & 98.73 & 54.70 & 9.22$\tcbhighmath[colback=LightOrange]{\downarrow 45.48}$ & 78.61 & 87.03 & 50.99 & 9.02$\tcbhighmath[colback=LightOrange]{\downarrow 41.97}$ & 79.06 & 84.90 & 56.68 & 12.14$\tcbhighmath[colback=LightOrange]{\downarrow 44.54}$ & 82.86 & 83.94 \\ 
        Qwen-VL-Chat  & 91.66 & 2.50$\tcbhighmath[colback=LightOrange]{\downarrow 89.16}$ & 97.35 & 97.35 & 62.32 & 14.45$\tcbhighmath[colback=LightOrange]{\downarrow 47.87}$ & 68.42 & 77.67 & 66.56 & 15.31$\tcbhighmath[colback=LightOrange]{\downarrow 51.25}$ & 71.77 & 77.61 & 88.01 & 46.66$\tcbhighmath[colback=LightOrange]{\downarrow 41.35}$ & 46.90 & 48.19   \\ 
        Otter  & 62.47 & 0.91$\tcbhighmath[colback=LightOrange]{\downarrow 61.56}$ & 98.42 & 98.91   & 53.55 & 18.71$\tcbhighmath[colback=LightOrange]{\downarrow 34.84}$ & 58.61 & 72.19 & 46.52 & 10.51$\tcbhighmath[colback=LightOrange]{\downarrow 36.01}$ & 66.90 & 81.49 & 45.68 & 26.02$\tcbhighmath[colback=LightOrange]{\downarrow 19.66}$ & 60.47 & 70.10 \\ 
        IDEFICS  & 44.28 & 1.29$\tcbhighmath[colback=LightOrange]{\downarrow 42.99}$ & 97.95 & 98.63 & 42.46 & 8.69$\tcbhighmath[colback=LightOrange]{\downarrow 33.77}$ & 75.57 & 87.47 & 32.70 & 5.46$\tcbhighmath[colback=LightOrange]{\downarrow 27.24}$ & 83.04 & 92.41 & 43.85 & 13.88$\tcbhighmath[colback=LightOrange]{\downarrow 29.97}$ & 81.31 & 85.64  \\ 
        InstructBLIP  & 76.04 & 1.06$\tcbhighmath[colback=LightOrange]{\downarrow 74.98}$ & 98.70 & 98.70 & 58.78 & 8.78$\tcbhighmath[colback=LightOrange]{\downarrow 50.00}$ & 77.68 & 87.33 & 54.72 & 5.30$\tcbhighmath[colback=LightOrange]{\downarrow 49.42}$ & 86.23 & 91.83 & 49.39 & 3.72$\tcbhighmath[colback=LightOrange]{\downarrow 45.67}$ & 94.62 & 95.39  \\ 
        \hline
        
   \small{\textit{13B LLM}} \\
        LLaVA-1.5-13B  & 81.27 & 1.67$\tcbhighmath[colback=LightOrange]{\downarrow 79.60}$ & 98.32 & 98.32 & 60.28 & 10.90$\tcbhighmath[colback=LightOrange]{\downarrow 49.38}$ & 77.21 & 86.18 & 57.70 & 6.87$\tcbhighmath[colback=LightOrange]{\downarrow 50.83}$ & 83.64 & 90.53 & 87.48 & 40.14$\tcbhighmath[colback=LightOrange]{\downarrow 47.34}$ & 57.24 & 57.33  \\ 
        ShareGPT4V-13B  & 84.31 & 1.36$\tcbhighmath[colback=LightOrange]{\downarrow 82.95}$ & 98.56 & 98.56 & 60.46 & 8.51$\tcbhighmath[colback=LightOrange]{\downarrow 51.95}$ & 78.74 & 88.86 & 58.53 & 7.04$\tcbhighmath[colback=LightOrange]{\downarrow 51.49}$ & 83.45 & 89.82 & 83.54 & 35.58$\tcbhighmath[colback=LightOrange]{\downarrow 47.96}$ & 59.85 & 60.22 \\ 
        InstructBLIP-13B  & 71.11 & 10.01$\tcbhighmath[colback=LightOrange]{\downarrow 61.10}$ & 85.18 & 89.66 & 49.29 & 9.40$\tcbhighmath[colback=LightOrange]{\downarrow 39.89}$ & 67.27 & 85.97 & 47.02 & 8.69$\tcbhighmath[colback=LightOrange]{\downarrow 38.33}$ & 71.30 & 87.15 & 45.52 & 22.08$\tcbhighmath[colback=LightOrange]{\downarrow 23.44}$ & 74.83 & 78.33\\ 
        \hline
        
   \small{\textit{Closed-Source}} \\
        GPT-4V & 90.14 & 85.52$\tcbhighmath[colback=LightOrange]{\downarrow 4.62}$ & 10.01 & 10.26  & 48.32 & 39.80$\tcbhighmath[colback=LightOrange]{\downarrow 8.52}$ & 21.83 & 31.56 & 41.64 & 34.19$\tcbhighmath[colback=LightOrange]{\downarrow 7.45}$ & 21.87 & 32.60 & 58.50 & 45.22$\tcbhighmath[colback=LightOrange]{\downarrow 13.28}$ & 45.78 & 48.64 \\
    
    \toprule[1.5pt]
    \end{tabular}}
    \caption{Detailed evaluation results for models answering the same questions under two different conversation settings: CleanConv. and HalluConv for each hallucination type. Numbers that are highlighted \tcbox[colback=LightOrange]{orange} represent the model performance drop caused by hallucinatory conversation, compared to the CleanConv.
    All experiments are implemented under a zero-shot setting to avoid the bias introduced by demonstrations. Note that all the models except GPT-4V have a significant drop in performance on HalluConv., and this trend holds for GPT-4V as well.}
    \label{moremoreresult}

\end{sidewaystable*}

\begin{table*}[t]
    \centering
    \resizebox{\linewidth}{!}{
    \begin{tabular}{lcccc|cccc}
    \toprule[1.5pt]
    \multirow{2}{*}{\textbf{Model}} 
    & \multicolumn{1}{c}{\textbf{CleanConv.}} & \multicolumn{3}{c}{\textbf{FactConv.}} & \multicolumn{1}{c}{\textbf{CleanConv.}} & \multicolumn{3}{c}{\textbf{IrrConv.}}\\
    \cmidrule(r){2-2} 
    \cmidrule(r){3-5} 
    \cmidrule(r){6-6}
    \cmidrule(r){7-9} 
   & \textbf{Acc}$\uparrow$ & \textbf{Acc}$\uparrow$ & \textbf{FR}$\downarrow$ & \multicolumn{1}{c}{\textbf{WFR}$\downarrow$} & \textbf{Acc}$\uparrow$ & \textbf{Acc}$\uparrow$ & \textbf{FR}$\downarrow$ & \multicolumn{1}{c}{\textbf{WFR}$\downarrow$}\\ \hline
   \small{\textit{7B LLM}} \\
        LLaVA-1.5 & 71.24 & 89.28$\tcbhighmath[colback=LightGreen]{\uparrow 18.04}$ & 4.74 & 6.35 & 71.24 & 65.35$\tcbhighmath[colback=LightOrange]{\downarrow 5.89}$ & 14.93 & 21.06\\ 
        MiniGPT-4 & 37.12 & 67.67$\tcbhighmath[colback=LightGreen]{\uparrow 30.55}$ & 5.85 & 9.15 & 37.12 & 35.11$\tcbhighmath[colback=LightOrange]{\downarrow 2.01}$ & 23.62 & 31.58 \\ 
        MiniGPT-v2 & 62.12 & 75.39$\tcbhighmath[colback=LightGreen]{\uparrow 13.27}$ & 11.01 & 14.83 & 62.12 & 56.46$\tcbhighmath[colback=LightOrange]{\downarrow 5.66}$ & 20.75 & 29.01 \\ 
        InternLM-XComposer & 43.51 & 74.04$\tcbhighmath[colback=LightGreen]{\uparrow 30.53}$ & 14.79 & 18.35 & 43.51 & 40.82$\tcbhighmath[colback=LightOrange]{\downarrow 2.69}$ & 25.83 & 40.34 \\ 
        ShareGPT4V & 71.81 & 89.32$\tcbhighmath[colback=LightGreen]{\uparrow 17.51}$ & 3.56 & 5.29 & 71.81 & 69.74$\tcbhighmath[colback=LightOrange]{\downarrow 2.07}$ & 10.47 & 16.27 \\  
        CogVLM  & 75.17 & 93.20$\tcbhighmath[colback=LightGreen]{\uparrow 18.03}$ & 0.64 & 1.82 & 75.17 & 74.68$\tcbhighmath[colback=LightOrange]{\downarrow 0.49}$ & 2.09 & 4.17 \\ 
        mPLUG-Owl  & 37.80 & 62.01$\tcbhighmath[colback=LightGreen]{\uparrow 24.21}$ & 13.56 & 27.23 & 37.80 & 30.54$\tcbhighmath[colback=LightOrange]{\downarrow 7.26}$ & 26.97 & 50.80 \\ 
        mPLUG-Owl2  & 60.47 & 88.06$\tcbhighmath[colback=LightGreen]{\uparrow 27.59}$ & 6.29 & 7.62 & 60.47 & 59.82$\tcbhighmath[colback=LightOrange]{\downarrow 0.65}$ & 18.99 & 25.01\\ 
        Qwen-VL-Chat  & 77.94 & 91.27$\tcbhighmath[colback=LightGreen]{\uparrow 13.33}$ & 0.72 & 1.68 & 77.94 & 77.12$\tcbhighmath[colback=LightOrange]{\downarrow 0.82}$ & 2.81 & 5.29 \\ 
        Otter  & 52.12 & 66.70$\tcbhighmath[colback=LightGreen]{\uparrow 14.58}$ & 14.62 & 22.45 & 52.12 & 44.06$\tcbhighmath[colback=LightOrange]{\downarrow 8.06}$ & 27.31 & 34.34 \\ 
        IDEFICS  & 40.94 & 73.68$\tcbhighmath[colback=LightGreen]{\uparrow 32.74}$ & 14.05 & 18.96 & 40.94 & 38.01$\tcbhighmath[colback=LightOrange]{\downarrow 2.93}$ & 29.96 & 47.64 \\ 
        InstructBLIP  & 59.88 & 86.10$\tcbhighmath[colback=LightGreen]{\uparrow 26.22}$ & 6.28 & 7.69  & 59.88 & 54.90$\tcbhighmath[colback=LightOrange]{\downarrow 4.98}$ & 17.90 & 23.47 \\ 
        \hline
        
   \small{\textit{13B LLM}} \\
        LLaVA-1.5-13B  & 72.43 & 90.85$\tcbhighmath[colback=LightGreen]{\uparrow 18.42}$ & 3.69 & 4.94 & 72.43 & 69.31$\tcbhighmath[colback=LightOrange]{\downarrow 3.12}$ & 11.22 & 17.49 \\ 
        ShareGPT4V-13B  & 72.43 & 91.98$\tcbhighmath[colback=LightGreen]{\uparrow 19.55}$ & 2.50 & 4.00 & 72.43 & 70.80$\tcbhighmath[colback=LightOrange]{\downarrow 1.63}$ & 9.69 & 15.82 \\ 
        InstructBLIP-13B  & 53.53 & 71.31$\tcbhighmath[colback=LightGreen]{\uparrow 17.78}$ & 16.27 & 22.69 & 53.53 & 42.71$\tcbhighmath[colback=LightOrange]{\downarrow 10.82}$ & 28.66 & 45.19 \\ 
        \hline
        
    \toprule[1.5pt]
    \end{tabular}}
    \caption{Experiment results for models answering the same questions under two different conversation settings: CleanConv., FactConv., IrrConv., and HalluConv. settings. Numbers that are highlighted \tcbox[colback=LightOrange]{orange} and \tcbox[colback=LightGreen]{green} represent the model performance drop and improvement in different conversation settings, compared to the model performance under CleanConv. setting.}
    \label{comparisonresult}
\end{table*}

\begin{table*}[h]
    \renewcommand{\arraystretch}{1.5}
    \centering
    \resizebox{\linewidth}{!}{
    \begin{tabular}{lllll|llll|llll|llll}
    \toprule[1.5pt]
    \multirow{3}{*}{\textbf{Model}} & \multicolumn{4}{c}{\textbf{Imagination}} & \multicolumn{4}{c}{\textbf{Existence}} & \multicolumn{4}{c}{\textbf{Attribute}} & \multicolumn{4}{c}{\textbf{Relation}} \\
    \cmidrule(r){2-5}
    \cmidrule(r){6-9}
    \cmidrule(r){10-13}
    \cmidrule(r){14-17}
    & \textbf{CleanConv.} & \multicolumn{3}{c}{\textbf{HalluConv.}} & \textbf{CleanConv.} & \multicolumn{3}{c}{\textbf{HalluConv.}} & \textbf{CleanConv.} & \multicolumn{3}{c}{\textbf{HalluConv.}} & \textbf{CleanConv.} & \multicolumn{3}{c}{\textbf{HalluConv.}}\\
    \cmidrule(r){2-2}
    \cmidrule(r){3-5}
    \cmidrule(r){6-6}
    \cmidrule(r){7-9}
    \cmidrule(r){10-10}
    \cmidrule(r){11-13}
    \cmidrule(r){14-14}
    \cmidrule(r){15-17}
   & \multicolumn{1}{c}{\textbf{Acc}$\uparrow$} & \multicolumn{1}{c}{\textbf{Acc}$\uparrow$} & \multicolumn{1}{c}{\textbf{FR}$\downarrow$} & \multicolumn{1}{c}{\textbf{WFR}$\downarrow$} & \multicolumn{1}{c}{\textbf{Acc}$\uparrow$} & \multicolumn{1}{c}{\textbf{Acc}$\uparrow$} & \multicolumn{1}{c}{\textbf{FR}$\downarrow$} & \multicolumn{1}{c}{\textbf{WFR}$\downarrow$} & \multicolumn{1}{c}{\textbf{Acc}$\uparrow$} & \multicolumn{1}{c}{\textbf{Acc}$\uparrow$} & \multicolumn{1}{c}{\textbf{FR}$\downarrow$} & \multicolumn{1}{c}{\textbf{WFR}$\downarrow$} & \multicolumn{1}{c}{\textbf{Acc}$\uparrow$} & \multicolumn{1}{c}{\textbf{Acc}$\uparrow$} & \multicolumn{1}{c}{\textbf{FR}$\downarrow$} & \multicolumn{1}{c}{\textbf{WFR}$\downarrow$}\\ \hline
        LLaVA-1.5 & 81.65 & 1.14 & 98.79 & 98.79  & 60.82 & 9.22 & 79.30 & 88.05 & 55.96 & 10.18 & 78.85 & 85.21 & 83.76 & 38.09 & 57.07 & 57.61 \\
        \qquad w/ \textit{Prompt} & 81.65$\tcbhighmath[colback=LightGreen]{\uparrow 0.00}$ & 1.36$\tcbhighmath[colback=LightGreen]{\uparrow 0.22}$ & 98.51$\tcbhighmath[colback=LightGreen]{\downarrow 0.28}$ & 98.51$\tcbhighmath[colback=LightGreen]{\downarrow 0.28}$ & 58.24$\tcbhighmath[colback=LightOrange]{\downarrow 2.58}$ & 9.13$\tcbhighmath[colback=LightOrange]{\downarrow 0.09}$ & 76.56$\tcbhighmath[colback=LightGreen]{\downarrow 2.74}$ & 87.67$\tcbhighmath[colback=LightGreen]{\downarrow 0.38}$ & 55.63$\tcbhighmath[colback=LightOrange]{\downarrow 0.33}$ & 6.87$\tcbhighmath[colback=LightOrange]{\downarrow 3.31}$ & 80.06$\tcbhighmath[colback=LightGreen]{\uparrow 1.21}$ & 89.58$\tcbhighmath[colback=LightGreen]{\uparrow 4.37}$ & 84.67$\tcbhighmath[colback=LightGreen]{\uparrow 0.91}$ & 35.13$\tcbhighmath[colback=LightOrange]{\downarrow 2.96}$ & 61.47$\tcbhighmath[colback=LightOrange]{\uparrow 4.40}$ & 62.01$\tcbhighmath[colback=LightOrange]{\uparrow 4.40}$ \\ 
        \qquad w/ \textit{VCD} & 76.27$\tcbhighmath[colback=LightOrange]{\downarrow 5.38}$ & 1.29$\tcbhighmath[colback=LightGreen]{\uparrow 0.15}$ & 98.61$\tcbhighmath[colback=LightGreen]{\downarrow 0.18}$ & 98.61$\tcbhighmath[colback=LightGreen]{\downarrow 0.18}$ & 60.55$\tcbhighmath[colback=LightOrange]{\downarrow 0.27}$ & 10.99$\tcbhighmath[colback=LightGreen]{\uparrow 1.77}$ & 76.87$\tcbhighmath[colback=LightGreen]{\downarrow 2.43}$ & 85.51$\tcbhighmath[colback=LightGreen]{\downarrow 2.54}$ & 55.96$\tcbhighmath[colback=LightGreen]{\uparrow 0.00}$ & 11.42$\tcbhighmath[colback=LightGreen]{\uparrow 1.24}$ & 76.92$\tcbhighmath[colback=LightGreen]{\downarrow 1.93}$ & 83.28$\tcbhighmath[colback=LightGreen]{\downarrow 1.93}$  & 85.43$\tcbhighmath[colback=LightGreen]{\uparrow 1.67}$ & 44.08$\tcbhighmath[colback=LightGreen]{\uparrow 5.99}$ & 50.36$\tcbhighmath[colback=LightGreen]{\downarrow 6.71}$ & 50.80$\tcbhighmath[colback=LightGreen]{\downarrow 6.81}$  \\ 
        \qquad w/ \textit{RVD (ours)} & 79.23$\tcbhighmath[colback=LightOrange]{\downarrow 2.42}$ & 12.13$\tcbhighmath[colback=LightGreen]{\uparrow 10.99}$ & 84.78$\tcbhighmath[colback=LightGreen]{\downarrow 14.01}$ & 84.78$\tcbhighmath[colback=LightGreen]{\downarrow 14.01}$ & 58.51$\tcbhighmath[colback=LightOrange]{\downarrow 2.31}$ & 37.94$\tcbhighmath[colback=LightGreen]{\uparrow 28.72}$ & 34.24$\tcbhighmath[colback=LightGreen]{\downarrow 45.06}$ & 45.61$\tcbhighmath[colback=LightGreen]{\downarrow 42.44}$ & 55.30$\tcbhighmath[colback=LightOrange]{\downarrow 0.66}$ & 24.01$\tcbhighmath[colback=LightGreen]{\uparrow 13.83}$ & 53.89$\tcbhighmath[colback=LightGreen]{\downarrow 24.96}$ & 66.92$\tcbhighmath[colback=LightGreen]{\downarrow 18.29}$ & 85.36$\tcbhighmath[colback=LightGreen]{\uparrow 1.60}$ & 57.28$\tcbhighmath[colback=LightGreen]{\uparrow 19.19}$ & 35.56$\tcbhighmath[colback=LightGreen]{\downarrow 21.51}$ & 36.00$\tcbhighmath[colback=LightGreen]{\downarrow 21.61}$  \\ \hline
        mPLUG-Owl2 & 77.86 & 1.21 & 98.73 & 98.73 & 54.70 & 9.22 & 78.61 & 87.03 & 50.99 & 9.02 & 79.06 & 84.90 & 56.68 & 12.14 & 82.86 & 83.94 \\ 
        \qquad w/ \textit{Prompt}  & 76.57$\tcbhighmath[colback=LightOrange]{\downarrow 1.29}$ & 1.21$\tcbhighmath[colback=LightGreen]{\uparrow 0.00}$ & 98.61$\tcbhighmath[colback=LightGreen]{\downarrow 0.12}$ & 98.61$\tcbhighmath[colback=LightGreen]{\downarrow 0.12}$ & 58.87$\tcbhighmath[colback=LightGreen]{\uparrow 4.17}$ & 7.89$\tcbhighmath[colback=LightOrange]{\downarrow 1.33}$ & 78.77$\tcbhighmath[colback=LightOrange]{\uparrow 0.16}$ & 88.10$\tcbhighmath[colback=LightOrange]{\uparrow 1.07}$  & 51.41$\tcbhighmath[colback=LightGreen]{\uparrow 0.42}$ & 9.44$\tcbhighmath[colback=LightGreen]{\uparrow 0.42}$ & 79.07$\tcbhighmath[colback=LightOrange]{\uparrow 0.01}$ & 84.86$\tcbhighmath[colback=LightGreen]{\downarrow 0.04}$ & 57.51$\tcbhighmath[colback=LightGreen]{\uparrow 0.83}$ & 12.75$\tcbhighmath[colback=LightGreen]{\uparrow 0.61}$ & 84.17$\tcbhighmath[colback=LightOrange]{\uparrow 1.31}$ & 84.70$\tcbhighmath[colback=LightOrange]{\uparrow 0.76}$ \\ 
        \qquad w/ \textit{VCD}  & 77.10$\tcbhighmath[colback=LightOrange]{\downarrow 0.76}$ & 1.14$\tcbhighmath[colback=LightOrange]{\downarrow 0.07}$ & 98.72$\tcbhighmath[colback=LightGreen]{\downarrow 0.01}$ & 98.72$\tcbhighmath[colback=LightGreen]{\downarrow 0.01}$ & 58.69$\tcbhighmath[colback=LightGreen]{\uparrow 3.99}$ & 10.02$\tcbhighmath[colback=LightGreen]{\uparrow 0.80}$ & 75.98$\tcbhighmath[colback=LightGreen]{\downarrow 2.63}$ & 85.80$\tcbhighmath[colback=LightGreen]{\downarrow 1.23}$ & 51.32$\tcbhighmath[colback=LightGreen]{\uparrow 0.33}$ & 11.92$\tcbhighmath[colback=LightGreen]{\uparrow 2.90}$ & 75.48$\tcbhighmath[colback=LightGreen]{\downarrow 3.58}$ & 81.45$\tcbhighmath[colback=LightGreen]{\downarrow 3.45}$ & 56.37$\tcbhighmath[colback=LightOrange]{\downarrow 0.31}$ & 12.44$\tcbhighmath[colback=LightGreen]{\uparrow 0.30}$ & 83.04$\tcbhighmath[colback=LightOrange]{\uparrow 0.18}$ & 83.85$\tcbhighmath[colback=LightGreen]{\downarrow 0.09}$ \\ 
        \qquad w/ \textit{RVD (ours)}  & 79.00$\tcbhighmath[colback=LightGreen]{\uparrow 1.14}$ & 16.22$\tcbhighmath[colback=LightGreen]{\uparrow 15.01}$ & 79.65$\tcbhighmath[colback=LightGreen]{\downarrow 19.08}$ & 79.65$\tcbhighmath[colback=LightGreen]{\downarrow 19.08}$  & 57.45$\tcbhighmath[colback=LightGreen]{\uparrow 2.75}$ & 37.59$\tcbhighmath[colback=LightGreen]{\uparrow 28.37}$ & 37.35$\tcbhighmath[colback=LightGreen]{\downarrow 41.26}$ & 48.77$\tcbhighmath[colback=LightGreen]{\downarrow 38.26}$ & 51.24$\tcbhighmath[colback=LightGreen]{\uparrow 0.25}$ & 23.59$\tcbhighmath[colback=LightGreen]{\uparrow 14.57}$ & 51.37$\tcbhighmath[colback=LightGreen]{\downarrow 27.69}$ & 63.17$\tcbhighmath[colback=LightGreen]{\downarrow 21.73}$ & 57.59$\tcbhighmath[colback=LightGreen]{\uparrow 0.91}$ & 15.02$\tcbhighmath[colback=LightGreen]{\uparrow 2.88}$ & 78.79$\tcbhighmath[colback=LightGreen]{\downarrow 4.07}$ & 79.58$\tcbhighmath[colback=LightGreen]{\downarrow 4.36}$ \\ \hline
        ShareGPT4V  & 84.08 & 1.82 & 98.20 & 98.20 & 59.49 & 10.99 & 76.30 & 84.20 & 56.79 & 9.52 & 79.30 & 86.30 & 83.84 & 40.06 & 55.29 & 55.66 \\  
        \qquad w/ \textit{Prompt}  & 85.52$\tcbhighmath[colback=LightGreen]{\uparrow 1.44}$ & 2.27$\tcbhighmath[colback=LightGreen]{\uparrow 0.45}$ & 97.70$\tcbhighmath[colback=LightGreen]{\downarrow 0.50}$ & 97.70$\tcbhighmath[colback=LightGreen]{\downarrow 0.50}$   & 57.09$\tcbhighmath[colback=LightOrange]{\downarrow 2.40}$ & 8.33$\tcbhighmath[colback=LightGreen]{\uparrow 2.66}$ & 78.42$\tcbhighmath[colback=LightOrange]{\uparrow 2.12}$ & 88.82$\tcbhighmath[colback=LightOrange]{\uparrow 4.62}$ & 56.95$\tcbhighmath[colback=LightGreen]{\uparrow 0.16}$ & 8.53$\tcbhighmath[colback=LightOrange]{\downarrow 0.99}$ & 80.96$\tcbhighmath[colback=LightOrange]{\uparrow 1.66}$ & 87.94$\tcbhighmath[colback=LightOrange]{\uparrow 1.64}$ & 83.84$\tcbhighmath[colback=LightGreen]{\uparrow 0.00}$ & 34.98$\tcbhighmath[colback=LightOrange]{\downarrow 5.08}$ & 61.00$\tcbhighmath[colback=LightOrange]{\uparrow 5.71}$ & 61.36$\tcbhighmath[colback=LightOrange]{\uparrow 5.70}$ \\ 
        \qquad w/ \textit{VCD}  & 81.20$\tcbhighmath[colback=LightOrange]{\downarrow 2.88}$ & 1.21$\tcbhighmath[colback=LightOrange]{\downarrow 0.61}$ & 98.79$\tcbhighmath[colback=LightOrange]{\uparrow 0.59}$ & 98.79$\tcbhighmath[colback=LightOrange]{\uparrow 0.59}$ & 63.12$\tcbhighmath[colback=LightGreen]{\uparrow 3.63}$ & 10.20$\tcbhighmath[colback=LightOrange]{\downarrow 0.79}$ & 77.53$\tcbhighmath[colback=LightOrange]{\uparrow 1.23}$ & 86.80$\tcbhighmath[colback=LightOrange]{\uparrow 2.60}$ & 61.01$\tcbhighmath[colback=LightGreen]{\uparrow 4.22}$ & 10.51$\tcbhighmath[colback=LightGreen]{\uparrow 0.99}$ & 77.48$\tcbhighmath[colback=LightGreen]{\downarrow 1.82}$ & 84.67$\tcbhighmath[colback=LightGreen]{\downarrow 1.63}$ & 83.92$\tcbhighmath[colback=LightGreen]{\uparrow 0.08}$ & 43.70$\tcbhighmath[colback=LightGreen]{\uparrow 3.64}$ & 50.54$\tcbhighmath[colback=LightGreen]{\downarrow 4.75}$ & 50.72$\tcbhighmath[colback=LightGreen]{\downarrow 4.94}$  \\ 
        \qquad w/ \textit{RVD (ours)}  & 86.05$\tcbhighmath[colback=LightGreen]{\uparrow 1.97}$ & 20.02$\tcbhighmath[colback=LightGreen]{\uparrow 18.20}$ & 77.53$\tcbhighmath[colback=LightGreen]{\downarrow 20.67}$ & 77.53$\tcbhighmath[colback=LightGreen]{\downarrow 20.67}$ & 57.45$\tcbhighmath[colback=LightOrange]{\downarrow 2.04}$ & 40.34$\tcbhighmath[colback=LightGreen]{\uparrow 29.35}$ & 29.48$\tcbhighmath[colback=LightGreen]{\downarrow 46.82}$ & 43.67$\tcbhighmath[colback=LightGreen]{\downarrow 40.53}$ & 58.61$\tcbhighmath[colback=LightGreen]{\uparrow 1.82}$ & 31.21$\tcbhighmath[colback=LightGreen]{\uparrow 21.69}$ & 45.06$\tcbhighmath[colback=LightGreen]{\downarrow 34.24}$ & 57.06$\tcbhighmath[colback=LightGreen]{\downarrow 29.24}$ & 83.46$\tcbhighmath[colback=LightOrange]{\downarrow 0.38}$ & 58.35$\tcbhighmath[colback=LightGreen]{\uparrow 18.29}$ & 32.91$\tcbhighmath[colback=LightGreen]{\downarrow 22.38}$ & 33.27$\tcbhighmath[colback=LightGreen]{\downarrow 22.39}$  \\ 
    
    \toprule[1.5pt]
    \end{tabular}}
    \caption{Detailed evaluation results for different methods on our proposed evaluation for each hallucination type. Numbers that are highlighted \tcbox[colback=LightOrange]{orange} and \tcbox[colback=LightGreen]{green} represent the model performance drop and improvement, respectively.}
    \label{rvdmoremoreresult}

\end{table*}

\end{document}